\documentclass[10pt,twocolumn,letterpaper]{article}

\usepackage[pagenumbers]{cvpr} 

\definecolor{cvprblue}{rgb}{0.21,0.49,0.74}
\usepackage[pagebackref,breaklinks,colorlinks,allcolors=cvprblue]{hyperref}
\usepackage{multirow}

\def\confName{CVPR}
\def\confYear{2027}

\title{Eq.Bot: Model-Agnostic Canonicalization for Equivariant Robotic Manipulation}

\author{
    Jian Deng\textsuperscript{1},
    Yuandong Wang\textsuperscript{1\dag},
    Yangfu Zhu\textsuperscript{1},
    Tao Feng\textsuperscript{2},
    Tianyu Wo\textsuperscript{3},
    Zhenzhou Shao\textsuperscript{1\dag}
    \\[4pt]
    \textsuperscript{1}Beijing Key Laboratory of Light Industrial Robot and Safety Verification, Capital Normal University \\
    \textsuperscript{2}Department of Computer Science and Technology, Tsinghua University \\
    \textsuperscript{3}School of Software, Beihang University \\
    \textsuperscript{\dag}Corresponding authors: {\tt\small wangyd@cnu.edu.cn, zshao@cnu.edu.cn}
}

\begin{document}
\maketitle
\begin{abstract}
Robotic manipulation is shifting from task-specific learning towards general-purpose vision-language-action (VLA) models. Despite their strong semantic generalization, these models remain sensitive to geometric transformations such as rotations. Existing approaches introduce equivariance through specialized architectures or symmetrization, often requiring network modifications or repeated policy evaluations. Meanwhile, heterogeneous action representations across VLA architectures further complicate maintaining geometric consistency. Inspired by human spatial reasoning, we propose Eq.Bot, a model-agnostic canonicalization framework grounded in group equivariance for geometrically consistent robotic manipulation. Eq.Bot aligns observations to a learned canonical frame, evaluates the base policy once, and recovers actions in the original coordinates through representation-specific mappings, preserving the policy's internal architecture. Experiments across four simulation benchmarks and three real-world tasks demonstrate improved manipulation performance across CNN- and Transformer-based policies of different scales while introducing modest inference overhead. The project website is available at: \textcolor{cyan}{https://dj9425.github.io/Eq.Bot}.
\end{abstract}    
\vspace{-1ex}
\section{Introduction}
\label{sec:intro}

Robotic manipulation is undergoing a shift from task-specific systems to general-purpose vision-language-action (VLA) models \cite{lynch2017modern, brohan2022rt, zitkovich2023rt, kim2024openvla, team2024octo, black2024pi_0}. By leveraging massive pre-training on internet-scale data, VLA models like OpenVLA-OFT \cite{kim2025fine} and $\pi_{0.5}$ \cite{intelligence2025pi_} have demonstrated remarkable semantic generalization capabilities. However, a critical "geometric gap" persists: semantic generalization does not ensure that predicted actions transform consistently with the observed scene. Recent evaluations show that VLA models remain substantially sensitive to variations in object layouts and camera viewpoints \cite{liu2025eva, li2026vla}. This limitation partly stems from a fundamental lack of geometric inductive bias: CNN-based architectures introduce aliasing through pooling \cite{zhang2019making}, while Transformer-based models break continuous symmetries via patch tokenization and absolute positional encodings \cite{dosovitskiy2020image, vaswani2017attention}.

\begin{figure}[t]
    \centering
    \includegraphics[width=1.0\linewidth]{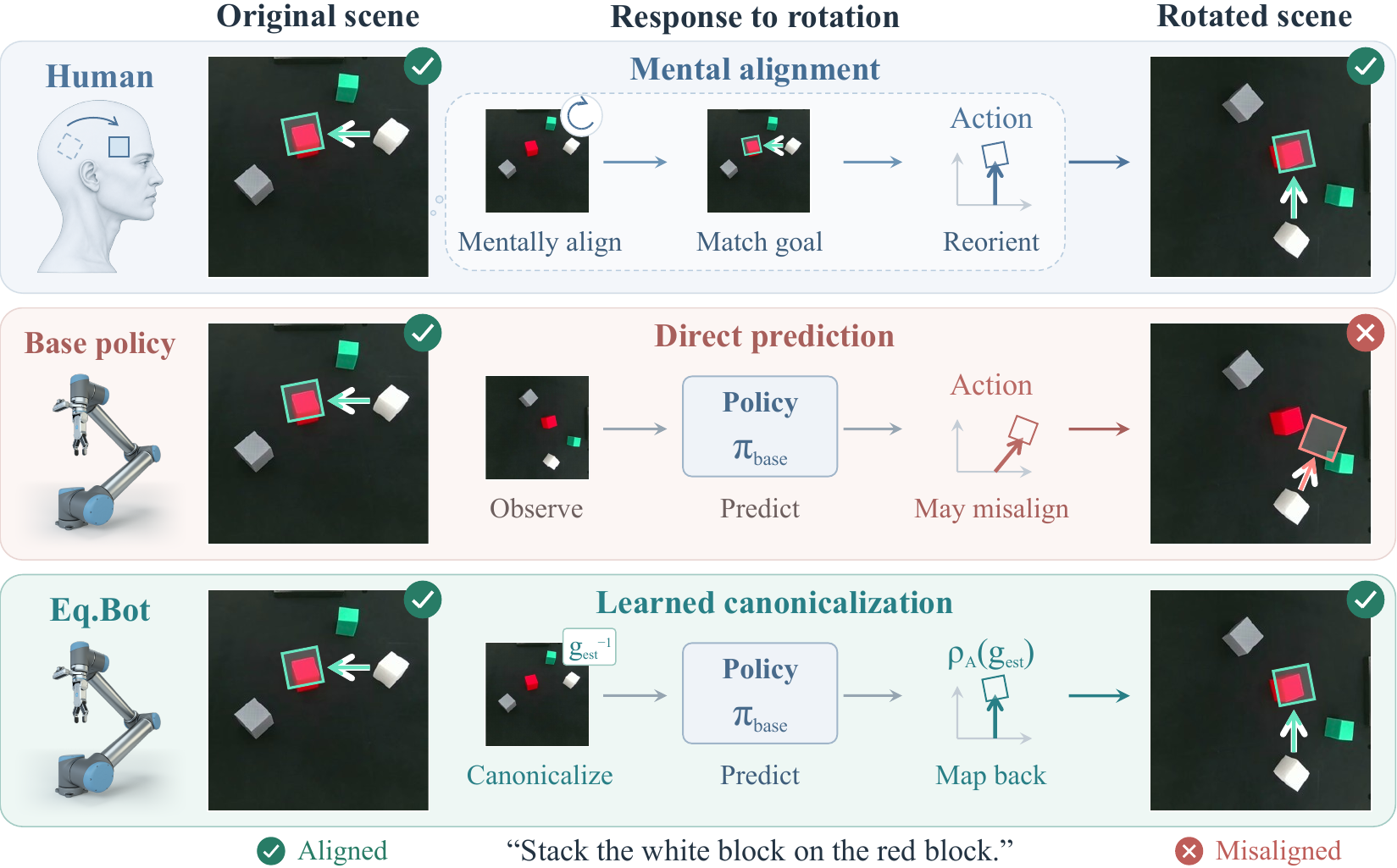}
    \caption{Intuition of Eq.Bot. Human mental shows adaptation to rotated observations, whereas direct policy prediction may yield geometric mismatches. Eq.Bot canonicalizes observations and maps predicted actions back to the original frame, enabling equivariant behavior without altering the base policy's architecture.}
    \label{fig:visual_intr}
    \vspace{-2ex}
\end{figure}

Existing works address this challenge through data augmentation, architectural constraints, or symmetrization. Data augmentation attempts to cover the transformation space through data-intensive training \cite{maharana2022review, dunlap2023diversify}. However, this approach relies on extensive training to statistically approximate symmetry and lacks theoretical constraints that guarantee generalization beyond the training distribution. Architectural constraints design bespoke equivariant architectures, such as group convolutions \cite{cohen2016group}, spherical harmonics \cite{hu20253d}, or steerable representations \cite{jia2025learning, chen2024eqvafford, tie2024seed, zhu2025equact}. Although theoretically feasible, these methods require modifications to the model's network architecture, making them difficult to apply to existing VLA models. Symmetrization methods, such as Frame Averaging, incorporate geometric symmetry by aggregating appropriately transformed predictions \cite{puny2021frame, kim2023learning}. Recent VLA methods, including EquiVLA and SAVLA, incorporate symmetry through specialized equivariant action heads while reusing pretrained vision-language backbones \cite{ha2026equivla, li2026savla}. Canonicalization offers an architecture-agnostic alternative, with established applications in visual perception, including image classification, point-cloud recognition, and segmentation \cite{kaba2023equivariance}. Extending this principle across manipulation policies requires coordinating observation alignment and action recovery while accommodating diverse output representations and preserving existing policy architectures.

Drawing inspiration from mental rotation in human spatial reasoning \cite{shepard1971mental}, we propose Eq.Bot, a model-agnostic canonicalization framework grounded in group equivariance for robotic manipulation. As illustrated in Figure~\ref{fig:visual_intr}, Eq.Bot aligns observations in a common frame and recovers actions in the original coordinates through mappings for diverse policy output representations. Observation canonicalization and action recovery operate outside the base policy, preserving its architecture. We instantiate two canonicalizers: DirNet estimates frames from group-equivariant features, while OptNet matches conventional encoder features of transformed views to a learnable reference.

Our evaluation spans Ravens, LIBERO, VLABench, and RoboTwin 2.0, together with three real-world manipulation tasks, covering CNN- and Transformer-based policies of different scales. The results demonstrate the effectiveness of our canonicalization framework across diverse policy architectures and task settings. Our contributions are summarized as follows:

1) We propose Eq.Bot, a model-agnostic framework that couples observation canonicalization with representation-specific action recovery while preserving the architectures of existing manipulation policies.

2) We instantiate DirNet and OptNet within a shared policy interface and formulate the conditions under which consistent frame estimation and compatible observation-action mappings yield an equivariant composite policy.

3) We demonstrate manipulation gains across CNN- and Transformer-based policies in simulation and real-world settings, with ablations on canonicalizer design and rotation groups and measurements of inference overhead.

\vspace{-0.5ex}
\section{Related Work}

\subsection{Vision-Language-Action Models}

Robotic manipulation has progressed from task-specific policies to general-purpose VLA models \cite{driess2023palm, ahn2022can, awadalla2023openflamingo}. Early works like CLIPORT \cite{shridhar2022cliport} build on Transporter \cite{zeng2021transporter}, combining pretrained vision-language features with convolutional spatial prediction. RT-1 \cite{brohan2022rt} scales Transformer-based policies with robot demonstrations, while RT-2 \cite{zitkovich2023rt} transfers web-pretrained vision-language knowledge to robotic control. Recently, OpenVLA \cite{kim2024openvla} and ${\pi}_0$ \cite{black2024pi_0} have further advanced this paradigm by fine-tuning massive pre-trained backbones (e.g., LLaMA, ViT), demonstrating impressive semantic generalization capabilities. However, despite their semantic prowess, these models do not inherently ensure that actions transform consistently with observations. As highlighted by recent evaluations \cite{liu2025eva, li2026vla}, VLA models are still sensitive to basic geometric transformations, such as object transformations and camera viewpoint changes.

\subsection{Equivariant Architectures for Robotic}

Equivariant architectures encode geometric symmetries directly in neural layers, as established by G-CNNs \cite{cohen2016group}. In robotics, Transporter Networks \cite{zeng2021transporter} leveraged this principle to achieve simple equivariance for pick-and-place tasks. Recent advances have extended this to more complex domains: GEM \cite{jia2025learning} incorporates language-conditioned $SE(2)$ equivariance, while EqvAfford \cite{chen2024eqvafford} uses $SE(3)$-equivariant representations for point-level affordance learning. ET-SEED \cite{tie2024seed} and EquAct \cite{zhu2025equact} extend $SE(3)$ equivariance to trajectory generation and language-conditioned multi-task manipulation, respectively. SAVLA \cite{li2026savla} couples observation canonicalization with an equivariant action head and geometric conditioning. While theoretically rigorous, these approaches typically require reconstructing the policy using specialized network architectures, limiting their transferability across different architectures.

\subsection{Symmetrization and Canonicalization}

Symmetrization provides an alternative to constraining backbone architectures. Frame Averaging \cite{puny2021frame} guarantees equivariance under suitable frame conditions, while Probabilistic Symmetrization \cite{kim2023learning} learns transformation distributions that yield equivariance in expectation. Canonicalization \cite{kaba2023equivariance} combines frame estimation with input-output transformations around non-equivariant backbones. In robotics, Canonical Policy \cite{zhang2025canonical} canonicalizes 3D point clouds and Cartesian actions. ISP \cite{hu20253d} and GEM \cite{jia2025learning} focus on specialized equivariant representations. GS-VLA \cite{park2026gs} uses depth and camera parameters to normalize viewpoints for frozen VLAs, addressing observation shifts without transforming policy outputs. These implementations target specific geometric representations or viewpoint adaptation. Extending canonicalization across existing policies requires recovery operators compatible with various action representations. Eq.Bot addresses this problem through a shared interface that couples learned image canonicalization with recovery of spatial, Cartesian, and joint-space outputs, while preserving the base policy's architecture.

\section{Preliminaries}

We define geometric equivariance for robotic manipulation and explain why standard policy architectures do not guarantee it, motivating Eq.Bot's canonicalization framework.

\noindent \textbf{Group Theory and Equivariance.}
A group $(G, \cdot)$ is a set $G$ with a binary operation $\cdot$, satisfying the axioms of closure, associativity, identity, and invertibility \cite{hamermesh2012group}. In this work, we consider a transformation group $G$, instantiated as $C_4\subset SO(2)\subset SE(2)$ in the main experiments, with rotations by multiples of $90^\circ$.

Let $\rho_{\mathcal{I}}(g)$ and $\rho_{\mathcal{A}}(g)$ denote the corresponding group actions on observations and policy outputs. A language-conditioned policy $\pi$ is \textit{equivariant} \cite{kerber2013applied} when:
\begin{equation}
    \pi(\rho_{\mathcal{I}}(g) \cdot I,\ell)
    =\rho_{\mathcal{A}}(g)\pi(I,\ell),
    \qquad \forall g\in G.
\end{equation}
Here, $I$ is the visual observation, $\ell$ is the instruction, and $A=\pi(I,\ell)$ is the predicted output. This property guarantees that a geometric transformation of the input results in a predictable, semantic-preserving transformation of the predicted action.

\noindent \textbf{Equivariance in Existing Policy Architectures.}
Standard CNN- and Transformer-based policies can learn geometric regularities from demonstrations, but their architectures do not generally enforce rotation equivariance. (See the appendix for a detailed analysis.)

1) CNN-based Architectures:
Standard convolutions are translation-equivariant under suitable boundary conditions, but their learned filters do not enforce rotation equivariance \cite{cohen2016group}. Downsampling through strided convolutions or pooling can additionally introduce aliasing and reduce shift consistency \cite{zhang2019making}. In CLIPORT \cite{shridhar2022cliport}, combining pretrained CLIP semantic features with a convolutional spatial stream does not by itself guarantee compatible transformation laws across the two streams. Consequently, semantic generalization and spatially structured prediction do not ensure an equivariant policy.

2) Transformer-based Architectures:
Standard visual transformers use patch embeddings and positional encodings that are not constrained to preserve rotational symmetry \cite{dosovitskiy2020image}. General rotations change both patch contents and their spatial arrangement; even rotations that preserve the patch grid need not induce equivariant token embeddings. Absolute positional encodings further anchor tokens to fixed grid locations without enforcing a compatible transformation law \cite{vaswani2017attention,chu2021conditional}. Self-attention therefore does not by itself guarantee rotation-equivariant action prediction.

These limitations motivate our model-agnostic canonicalization framework that introduces geometric consistency at the observation-action interface, without modifying the model's network architecture.

\section{Method} \label{method}

Given demonstrations $\mathcal{D}=\{(I_i,\ell_i,A_i)\}_{i=1}^{N}$, where $I_i$ denotes the visual observation, $\ell_i$ the language instruction, and $A_i$ the demonstrated action or action chunk, we seek a manipulation policy $\pi: (I, \ell) \rightarrow A$ whose predictions remain geometrically consistent under transformations in group $G$. Formally, this requires the policy to satisfy the condition $\pi(g \cdot I, \ell) = g \cdot \pi(I, \ell)$, ensuring that the predicted action consistently tracks the object's pose changes. However, standard VLA architectures do not inherently guarantee this geometric equivariance. Consequently, our goal is to develop a universal framework that couples learned observation canonicalization with representation-specific action recovery to obtain this property under consistent frame estimation and compatible mappings, while preserving the base policy's architecture.
\begin{figure*}[tp]
    \centering
    \includegraphics[width=1.0\linewidth]{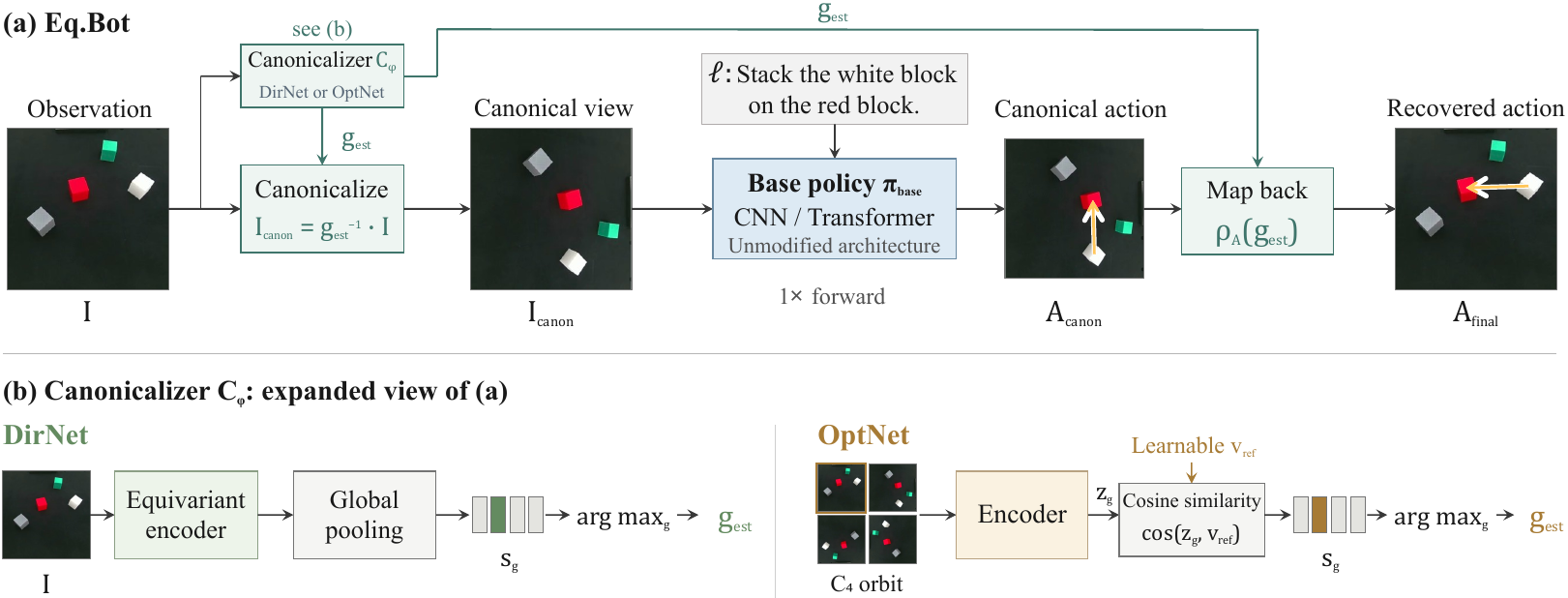}
    \caption{Overview of Eq.Bot. (a) A learned canonicalizer estimates a frame transform, aligns the observation, and reuses the same transform to map the base policy's prediction back to the original coordinates. The base policy retains its architecture and is forward once. (b) DirNet scores candidate frames through group-equivariant features, whereas OptNet matches orbit features to a learnable reference. The two canonicalizers are alternative implementations optimized jointly with the trainable policy components.}
    \label{fig:overview}
\vspace{-2ex}
\end{figure*}

\subsection{The Eq.Bot Framework}

As illustrated in Figure~\ref{fig:overview}(a), Eq.Bot wraps an existing policy with three stages: canonicalize the observation, predict an action in the canonical frame, and map the prediction back to the original coordinates. The estimated transformation connects the input and output mappings, allowing the base policy to operate in a common geometric frame.

\noindent \textbf{Phase I: Observation Canonicalization.}
For an observation $I$, a canonicalizer $C_{\phi}$ estimates a frame transformation $g_{\text{est}}\in G$. We define $g_{\text{est}}$ as the transformation from the canonical frame to the original frame, so the observation is canonicalized using its inverse:
\begin{equation}
    g_{\text{est}}=C_{\phi}(I), \qquad
    I_{\text{canon}}=g_{\text{est}}^{-1}\cdot I,
\end{equation}
where $g\cdot I$ denotes the group action on observations. DirNet and OptNet provide two implementations of $C_{\phi}$, detailed in Section~\ref{sec:canonicalization}. When frame estimates are consistent with the group action, transformed observations share the same canonical view.

\noindent \textbf{Phase II: Base Policy Inference.}
The base policy receives the canonical observation $I_{\text{canon}}$ and the unchanged language instruction $\ell$:
\begin{equation}
    A_{\text{canon}}=\pi_{\text{base}}(I_{\text{canon}},\ell).
\end{equation}
This stage preserves the policy architecture and its action representation, accommodating both CNN-based spatial prediction and Transformer-based action generation. The base policy processes only the selected canonical view, independently of the number of candidate frames considered by the canonicalizer.

\noindent \textbf{Phase III: Action Inverse Transformation.}
Finally, the same estimated frame is reused to recover an action in the original coordinates:
\begin{equation}
    A_{\text{final}}=\rho_{\mathcal{A}}(g_{\text{est}})A_{\text{canon}},
    \label{eq:action_mapping}
\end{equation}
where $\rho_{\mathcal{A}}$ is the corresponding group action on the policy's output representation. This mapping reverses observation canonicalization and links the policy prediction to the original scene geometry.

\subsection{Group-Equivariant Canonicalization} \label{sec:canonicalization}

The canonicalizer learns a reference frame suited to the manipulation objective. We focus on the group $C_4$, applying transformations to the entire scene image without object-specific segmentation. This design is robust to background complexity, as rotating the global observation also transforms the target object, enabling reliable localization without explicit object segmentation. Figure~\ref{fig:overview}(b) presents two alternative frame-scoring strategies: direct canonicalization with equivariant features and optimization-based canonicalization with conventional encoder features.

\noindent \textbf{Orbit and Frame Selection.}
The candidate canonical views form the orbit $\mathcal{T}(I)=\{I_g=g^{-1}\cdot I\mid g\in G\}$. Indexing candidates by $g^{-1}$ ensures that the selected $g$ can be reused directly for action recovery in Eq.~\ref{eq:action_mapping}. Both variants assign a score $s_g(I)$ to each candidate frame and select:
\begin{equation}
    g_{\text{est}}=\underset{g\in G}{\arg\max}\;s_g(I).
\end{equation}

\noindent \textbf{Direct Canonicalization (DirNet).}
DirNet employs G-CNNs based on the E2CNN \cite{weiler2019general}, to extract features equivariant to the selected rotation group. Steerable convolutions encode the candidate orientations in a group feature dimension. Global pooling over spatial locations and feature channels, while retaining this group dimension, yields the frame scores:
\begin{equation}
    s_g(I)=\big[\operatorname{GlobalPool}(f_{\text{eq}}(I))\big]_g.
\end{equation}
Thus, DirNet obtains orientation scores directly from the observation through an explicit geometric inductive bias. When the group scores transform equivariantly and the maximizing frame is unique, frame selection is consistent with canonicalization rotations.

\noindent \textbf{Optimization-based Canonicalization (OptNet).}
To prioritize training efficiency and computational overhead, OptNet learns canonicalization using conventional networks, such as CNNs or ResNet-50. A shared non-equivariant encoder extracts $\mathbf{z}_g=f_{\text{non-eq}}(I_g)$ from each orbit view. These features are scored against a learnable reference vector $\mathbf{v}_{\text{ref}}$ by cosine similarity:
\begin{equation}
    s_g(I)=\frac{\mathbf{z}_g^{\top}\mathbf{v}_{\text{ref}}}
    {\|\mathbf{z}_g\|\,\|\mathbf{v}_{\text{ref}}\|}.
\end{equation}
The reference is learned from task supervision, allowing frame selection to adapt to the downstream manipulation objective without requiring an equivariant encoder.

\noindent \textbf{Joint Optimization.}
We jointly optimize the canonicalizer parameters $\phi$ and the trainable policy parameters $\theta$ under task supervision; for OptNet, $\phi$ also includes $\mathbf{v}_{\text{ref}}$. Writing $\pi_{\text{Eq.Bot}}^{\theta,\phi}$ for the composite policy, the task objective is:
\begin{equation}
    \min_{\theta,\phi}\;
    \mathbb{E}_{(I,\ell,A_{\text{gt}})\sim\mathcal{D}}
    \left[\mathcal{L}_{\text{task}}
    \big(\pi_{\text{Eq.Bot}}^{\theta,\phi};I,\ell,A_{\text{gt}}\big)\right],
\end{equation}
where $\mathcal{L}_{\text{task}}$ denotes the supervised objective associated with the base policy. Discrete frame selection uses a hard choice in the forward pass and a straight-through softmax estimator in the backward pass. This enables task supervision to guide frame selection while retaining a single canonical view for policy inference.

\subsection{Action Inverse Transformation}

The inverse transformation maps the actions predicted in the canonical frame back to the original workspace coordinates, which is essential to reconstruct the correct physical action and support the overall equivariance of the pipeline under compatible observation-action transformations.

The action mapping in Eq.~\ref{eq:action_mapping} is defined according to the base policy's output representation. For pixel-wise affordance maps, as in CLIPORT, $\rho_{\mathcal{A}}(g_{\text{est}})$ rotates the predicted spatial grid back to the original observation coordinates, recovering the corresponding pick-and-place locations while preserving relative placement angles.

For continuous end-effector actions in physical coordinates, the inverse transformation composes absolute poses with the estimated frame and re-encodes orientations in the policy's representation. For world-frame increments, it rotates translation and rotation vectors without adding the positional offsets. For joint-space actions, as in TurboVLA, it applies forward kinematics, the workspace transformation, and inverse kinematics on a consistent reachable branch. Gripper commands remain unchanged. These representation-specific mappings allow Eq.Bot to retain the base policy's action interface while maintaining a consistent relation between observation and action coordinates. Implementation details are provided in the appendix.

\subsection{Equivariance Analysis}

We establish the condition under which the observation and action mappings yield an equivariant policy. Assume that the estimated frame satisfies:
\begin{equation}
    g_{\text{est}}(h\cdot I)=h\,g_{\text{est}}(I),
    \qquad h\in G,
\end{equation}
and that the observation and action transformations are compatible group actions on the valid input-output domain. The canonical observation is then invariant:
\begin{equation}
\begin{aligned}
    I_{\text{canon}}(h\cdot I)
    &=\big(h\,g_{\text{est}}(I)\big)^{-1}\cdot(h\cdot I)\\
    &=I_{\text{canon}}(I).
\end{aligned}
\end{equation}
With the language instruction held fixed, the composite policy therefore satisfies:
\begin{equation}
\begin{aligned}
    &\pi_{\text{Eq.Bot}}(h\cdot I,\ell) \\
    &= \rho_{\mathcal{A}}\big(g_{\text{est}}(h\cdot I)\big)
       \pi_{\text{base}}\big(I_{\text{canon}}(h\cdot I),\ell\big) \\
    &= \rho_{\mathcal{A}}\big(h\,g_{\text{est}}(I)\big)
       \pi_{\text{base}}\big(I_{\text{canon}}(I),\ell\big) \\
    &= \rho_{\mathcal{A}}(h)
       \Big[
       \rho_{\mathcal{A}}\big(g_{\text{est}}(I)\big)
       \pi_{\text{base}}\big(I_{\text{canon}}(I),\ell\big)
       \Big] \\
    &= \rho_{\mathcal{A}}(h)\,
       \pi_{\text{Eq.Bot}}(I,\ell).
\end{aligned}
\end{equation}
This derivation establishes the equivariance of $\pi_{\text{Eq.Bot}}$ under consistent frame estimation and compatible observation–action transformations. In practice, frame ambiguities and numerical resampling may affect these conditions. Joint optimization encourages task-relevant frame selection, while inverse action mapping transforms predictions back to the original coordinates. Together, these mechanisms promote geometric consistency while preserving the underlying policy architecture.
\section{Experiments}

\begin{table*}[ht] 
    \caption{Comparison of CNN-based policies on Ravens tasks. Task completion scores (\%) are reported across tasks and demonstration budgets under \textcolor{blue}{seen} and \textcolor{red}{unseen} settings. Bold and underlining indicate the best and second-best results, respectively.} 
    \centering 
    \setlength{\tabcolsep}{3.6pt}
    \resizebox{\textwidth}{!}{ 
        \begin{tabular}{lcccccccccccccccccccccccc} 
        \toprule
        & \multicolumn{4}{c}{packing-box-pairs} & \multicolumn{4}{c}{packing-box-pairs} & \multicolumn{4}{c}{packing-\textcolor{blue}{seen}-google} & \multicolumn{4}{c}{packing-\textcolor{red}{unseen}-google} & \multicolumn{4}{c}{packing-\textcolor{blue}{seen}-google} & \multicolumn{4}{c}{packing-\textcolor{red}{unseen}-google} \\ 
        & \multicolumn{4}{c}{\textcolor{blue}{seen}-colors} & \multicolumn{4}{c}{\textcolor{red}{unseen}-colors} & \multicolumn{4}{c}{objects-seq} & \multicolumn{4}{c}{objects-seq} & \multicolumn{4}{c}{objects-group} & \multicolumn{4}{c}{objects-group} \\[-4pt] 
        & \multicolumn{4}{r}{\rule[2pt]{3.6cm}{0.6pt}} & \multicolumn{4}{r}{\rule[2pt]{3.6cm}{0.6pt}} & \multicolumn{4}{r}{\rule[2pt]{3.6cm}{0.6pt}} & \multicolumn{4}{r}{\rule[2pt]{3.6cm}{0.6pt}} & \multicolumn{4}{r}{\rule[2pt]{3.6cm}{0.6pt}} & \multicolumn{4}{r}{\rule[2pt]{3.6cm}{0.6pt}} \\[-2pt] 
        Method & 1 & 10 & 100 & 1000 & 1 & 10 & 100 & 1000 & 1 & 10 & 100 & 1000 & 1 & 10 & 100 & 1000 & 1 & 10 & 100 & 1000 & 1 & 10 & 100 & 1000 \\ 
        \midrule 
        Transporter-only & 44.2 & 55.2 & 54.2 & 52.4 & 34.6 & 48.7 & 47.2 & 54.1 & 26.2 & 39.7 & 45.4 & 46.3 & 19.9 & 29.8 & 28.7 & 37.3 & \textbf{60.0} & 54.3 & 61.5 & 59.9 & 46.2 & 54.7 & 49.8 & 52.0 \\ 
        CLIP-only & 38.6 & 69.7 & 88.5 & 87.1 & 33.0 & 65.5 & 68.8 & 61.2 & \underline{29.1} & \textbf{67.9} & \underline{89.3} & \underline{95.8} & \textbf{37.1} & \textbf{49.4} & 60.4 & 57.8 & \underline{52.5} & 62.0 & \underline{89.6} & 92.7 & 43.4 & \textbf{65.9} & 73.1 & 70.0 \\ 
        RN50-BERT & 36.2 & 64.0 & \underline{94.7} & 90.3 & 31.4 & 52.7 & 65.6 & 72.1 & \textbf{32.9} & 48.4 & 87.9 & 94.0 & 29.3 & \underline{48.5} & 48.3 & 56.1 & 46.4 & 52.9 & 76.5 & 86.4 & 43.2 & 52.0 & 66.3 & 73.7 \\
        CLIPORT & 46.1 & 71.2 & 88.5 & 97.6 & 38.1 & 54.7 & 64.5 & 71.5 & 11.2 & 51.0 & 80.3 & 93.2 & 23.4 & 44.3 & 59.6 & 61.6 & 45.6 & 60.7 & 81.2 & 92.9 & 47.9 & 57.3 & 72.0 & 72.6 \\
        \midrule 
        Eq.Bot (OptNet) & \underline{53.9} & \underline{75.1} & 94.5 & \underline{98.9} & \textbf{47.1} & \underline{69.0} & \underline{74.7} & \underline{77.2} & 15.8 & 57.5 & 89.2 & \textbf{97.3} & \underline{33.5} & 45.9 & \underline{62.9} & \underline{73.2} & 49.3 & \underline{65.2} & 88.2 & \textbf{97.6} & \underline{50.1} & 59.8 & \underline{80.6} & \underline{77.3} \\
        Eq.Bot (DirNet) & \textbf{54.3} & \textbf{76.5} & \textbf{95.2} & \textbf{99.0} & \underline{46.7} & \textbf{71.8} & \textbf{78.1} & \textbf{78.8} & 16.4 & \underline{58.2} & \textbf{89.4} & \textbf{97.3} & 32.1 & 47.3 & \textbf{63.3} & \textbf{73.5} & 50.2 & \textbf{67.1} & \textbf{90.1} & \underline{97.5} & \textbf{51.3} & \underline{60.3} & \textbf{80.9} & \textbf{79.2} \\
        \midrule 
        & \multicolumn{4}{c}{stack-block-pyramid} & \multicolumn{4}{c}{stack-block-pyramid} & \multicolumn{4}{c}{separating-piles} & \multicolumn{4}{c}{separating-piles} & \multicolumn{4}{c}{towers-of-hanoi} & \multicolumn{4}{c}{towers-of-hanoi} \\ 
        & \multicolumn{4}{c}{seq-\textcolor{blue}{seen}-colors} & \multicolumn{4}{c}{seq-\textcolor{red}{unseen}-colors} & \multicolumn{4}{c}{\textcolor{blue}{seen}-colors} & \multicolumn{4}{c}{\textcolor{red}{unseen}-colors} & \multicolumn{4}{c}{seq-\textcolor{blue}{seen}-colors} & \multicolumn{4}{c}{seq-\textcolor{red}{unseen}-colors} \\[-4pt] 
        & \multicolumn{4}{r}{\rule[2pt]{3.6cm}{0.6pt}} & \multicolumn{4}{r}{\rule[2pt]{3.6cm}{0.6pt}} & \multicolumn{4}{r}{\rule[2pt]{3.6cm}{0.6pt}} & \multicolumn{4}{r}{\rule[2pt]{3.6cm}{0.6pt}} & \multicolumn{4}{r}{\rule[2pt]{3.6cm}{0.6pt}} & \multicolumn{4}{r}{\rule[2pt]{3.6cm}{0.6pt}} \\[-2pt] 
        Method & 1 & 10 & 100 & 1000 & 1 & 10 & 100 & 1000 & 1 & 10 & 100 & 1000 & 1 & 10 & 100 & 1000 & 1 & 10 & 100 & 1000 & 1 & 10 & 100 & 1000 \\ 
        \midrule 
        Transporter-only & 4.5 & 2.3 & 5.2 & 4.5 & 3.0 & 4.0 & 2.3 & 5.8 & 42.7 & 52.3 & 42.0 & 48.4 & 41.2 & 49.2 & 44.7 & 52.3 & 25.4 & 67.9 & \underline{98.0} & \underline{99.9} & 24.3 & 44.6 & 71.7 & 80.7 \\
        CLIP-only & 6.3 & 28.7 & 55.7 & 54.8 & 2.0 & 12.2 & 18.3 & 19.5 & 43.5 & 55.0 & 84.9 & 90.2 & \textbf{59.9} & 49.6 & 73.0 & 71.0 & 9.4 & 52.6 & 88.6 & 45.3 & 24.7 & 47.0 & 67.0 & 58.0 \\
        RN50-BERT & 5.3 & 35.0 & 89.0 & 97.5 & 6.2 & 12.2 & 21.5 & \underline{30.7} & 31.8 & 47.8 & 46.5 & 46.5 & 33.4 & 44.4 & 41.3 & 44.9 & 28.0 & 66.1 & 91.3 & 92.1 & 17.4 & 75.1 & 85.3 & 89.3 \\
        CLIPORT & 17.4 & 44.7 & 87.9 & 96.8 & 8.9 & 21.7 & 29.5 & 26.5 & 45.8 & 53.5 & 86.0 & 96.5 & 42.3 & 49.2 & 72.1 & 74.2 & 47.4 & 74.9 & 96.7 & 98.6 & \underline{43.3} & \underline{84.6} & 94.0 & \underline{99.1} \\
        \midrule 
        Eq.Bot (OptNet) & \underline{24.0} & \underline{53.7} & \underline{94.2} & \textbf{99.5} & \underline{17.3} & \textbf{30.5} & \underline{33.0} & \textbf{31.5} & \textbf{53.2} & \underline{60.9} & \textbf{94.1} & \underline{98.1} & 46.1 & \underline{58.2} & \underline{75.6} & \textbf{79.6} & \underline{56.2} & \textbf{86.0} & \textbf{100} & \textbf{100} & 42.7 & \textbf{84.7} & \textbf{99.9} & \textbf{100} \\
        Eq.Bot (DirNet) & \textbf{24.6} & \textbf{54.2} & \textbf{95.0} & \underline{99.0} & \textbf{18.5} & \underline{28.5} & \textbf{34.0} & 29.8 & \underline{52.8} & \textbf{61.5} & \underline{93.2} & \textbf{98.3} & \underline{47.3} & \textbf{58.9} & \textbf{77.1} & \underline{79.3} & \textbf{57.6} & \underline{85.3} & \textbf{100} & \textbf{100} & \textbf{43.8} & 83.2 & \underline{99.7} & \textbf{100} \\
        \midrule 
        & \multicolumn{4}{c}{\multirow{2}{*}{align-rope}} & \multicolumn{4}{c}{\multirow{2}{*}{packing-\textcolor{red}{unseen}-shapes}} & \multicolumn{4}{c}{assembling-kits-seq} & \multicolumn{4}{c}{assembling-kits-seq} & \multicolumn{4}{c}{put-blocks-in-bowls} & \multicolumn{4}{c}{put-blocks-in-bowls} \\ 
        & \multicolumn{4}{c}{} & \multicolumn{4}{c}{} & \multicolumn{4}{c}{\textcolor{blue}{seen}-colors} & \multicolumn{4}{c}{\textcolor{red}{unseen}-colors} & \multicolumn{4}{c}{\textcolor{blue}{seen}-colors} & \multicolumn{4}{c}{\textcolor{red}{unseen}-colors} \\[-4pt] 
        & \multicolumn{4}{r}{\rule[2pt]{3.6cm}{0.6pt}} & \multicolumn{4}{r}{\rule[2pt]{3.6cm}{0.6pt}} & \multicolumn{4}{r}{\rule[2pt]{3.6cm}{0.6pt}} & \multicolumn{4}{r}{\rule[2pt]{3.6cm}{0.6pt}} & \multicolumn{4}{r}{\rule[2pt]{3.6cm}{0.6pt}} & \multicolumn{4}{r}{\rule[2pt]{3.6cm}{0.6pt}} \\[-2pt] 
        Method & 1 & 10 & 100 & 1000 & 1 & 10 & 100 & 1000 & 1 & 10 & 100 & 1000 & 1 & 10 & 100 & 1000 & 1 & 10 & 100 & 1000 & 1 & 10 & 100 & 1000 \\ 
        \midrule 
        Transporter-only & 6.9 & 30.6 & 33.1 & 51.5 & 16.0 & 20.0 & 22.0 & 22.0 & 5.8 & 11.6 & 28.6 & 29.6 & 7.8 & 17.6 & 25.6 & 28.4 & 16.8 & 33.3 & 62.7 & 64.7 & 11.7 & 17.2 & 14.8 & 18.7 \\
        CLIP-only & 13.4 & 48.7 & 70.4 & 70.7 & 13.0 & 28.0 & 44.0 & 50.0 & 0.8 & 9.2 & 19.8 & 23.0 & 2.0 & 4.6 & 10.8 & 19.8 & 23.5 & \textbf{60.2} & 93.5 & \underline{97.7} & 11.2 & \textbf{34.2} & 33.2 & 44.5 \\
        RN50-BERT & 3.1 & 25.0 & 63.8 & 57.1 & 19.0 & 25.0 & 32.0 & 44.0 & 2.2 & 5.6 & 11.6 & 21.8 & 1.6 & 6.4 & 10.4 & 18.4 & 13.8 & 44.5 & 81.2 & 91.8 & 16.2 & 23.0 & 30.3 & 23.8 \\
        CLIPORT & 15.9 & 48.6 & 62.4 & 90.6 & 22.0 & 28.0 & 32.0 & 38.0 & 9.8 & 14.2 & 36.8 & 53.2 & 12.6 & 17.4 & \textbf{33.6} & \underline{31.0} & 20.4 & 38.3 & 91.2 & \textbf{100} & 16.0 & 29.0 & 37.1 & 18.5 \\
        \midrule 
        Eq.Bot (OptNet) & \underline{23.5} & \underline{62.6} & \underline{92.6} & \underline{94.5} & \textbf{25.0} & \underline{38.0} & \underline{50.0} & \underline{54.0} & \textbf{13.6} & \underline{18.3} & \underline{48.2} & \underline{53.6} & \textbf{18.2} & \underline{21.6} & \underline{32.0} & 30.4 & \underline{31.5} & 52.7 & \underline{99.5} & \textbf{100} & \underline{19.5} & \underline{32.7} & \textbf{59.8} & \textbf{52.3} \\
        Eq.Bot (DirNet) & \textbf{24.1} & \textbf{64.9} & \textbf{93.6} & \textbf{96.2} & \underline{24.0} & \textbf{44.0} & \textbf{51.0} & \textbf{55.0} & \underline{13.2} & \textbf{19.4} & \textbf{51.6} & \textbf{54.8} & \underline{17.8} & \textbf{22.4} & 31.0 & \textbf{33.0} & \textbf{32.7} & \underline{53.2} & \textbf{99.8} & \textbf{100} & \textbf{20.3} & 32.5 & \underline{58.5} & \underline{52.2} \\
        \bottomrule
        \end{tabular}} 
    \label{tab:CLIPORT_label}
    \vspace{-1ex}
\end{table*}

We evaluate the effectiveness of Eq.Bot's canonicalization across CNN- and Transformer-based policies, as well as its inference overhead. Experiments span four simulation benchmarks—Ravens, LIBERO, VLABench, and RoboTwin 2.0—to three real-world tasks, covering diverse policy architectures and task settings. Ablations examine the effects of canonicalization networks and rotation groups, while latency quantifies computational cost.


 




\subsection{Experimental Setup}

This section summarizes the data sources, implementation details, and training configurations for our experiments, with further details provided in the appendix.

\noindent \textbf{Data Sources.}
We use public datasets from Ravens \cite{shridhar2022cliport}, LIBERO \cite{liu2023libero}, VLABench \cite{zhang2025vlabench}, and RoboTwin 2.0 \cite{chen2025robotwin}. Ravens uses 1, 10, 100, or 1000 demonstrations for each of 18 task settings, and LIBERO uses the modified RLDS datasets released with OpenVLA \cite{kim2024openvla}. VLABench provides 500 demonstrations for each of 10 training tasks, totaling 5,000 trajectories. RoboTwin 2.0 uses 50 Clean and 500 Randomized demonstrations per task across 50 tasks. We additionally collect 50 teleoperated demonstrations for each of \textit{Place Object}, \textit{Pack Box}, and \textit{Stack Blocks}, totaling 150 real-world trajectories.

\noindent \textbf{Implementation Details.}
We build on the released implementations of CLIPORT, OpenVLA-OFT, $\pi_{0.5}$, and TurboVLA \cite{shridhar2022cliport,kim2025fine,intelligence2025pi_,xie2026turbovla}, retaining their internal architectures, observation preprocessing, and native action representations. Eq.Bot uses a shared estimated frame for observation canonicalization and representation-specific action recovery. Unless otherwise stated, we use $C_4$ rotations, with DirNet based on group-equivariant convolutions and OptNet on a ResNet-50 encoder. Real-world experiments use a UR5e arm equipped with a Robotiq 2F-85 gripper and two Intel RealSense cameras providing external and wrist-mounted RGB views. Deployment runs on a server with two NVIDIA RTX 2080 Ti GPUs.

\noindent \textbf{Training Configuration.}
We follow the official training recipes of each policy, adapting batch sizes and distributed training to the available GPU resources. The original objectives, freezing schemes, and adaptation strategies are retained, and canonicalizers are jointly optimized with the trainable policy components. Training runs 200,000 for CLIPORT (effective batch size 1), 150,000 for OpenVLA-OFT and TurboVLA (batch sizes 4 and 8), and 60,000 for $\pi_{0.5}$ (batch size 64). For the CNN-based CLIPORT policy, we use learning rates of $10^{-4}$ for the policy and $10^{-3}$ for the canonicalizer to facilitate faster adaptation of the latter. For Transformer-based policies, we retain their respective learning-rate schedules, enabling the canonicalizer to adapt alongside the trainable components. Real-world policies follow the OpenVLA-OFT fine-tuning protocol.

\subsection{Simulation Experiments}

We compare each base policy with its OptNet and DirNet variants across policy architectures, action generation paradigms, and model scales. For CNN-based policies, we evaluate CLIPORT on Ravens, testing spatial manipulation across demonstration budgets and seen or unseen object attributes. For Transformer-based policies, we evaluate large-scale OpenVLA-OFT with direct action regression on LIBERO suites, flow matching based $\pi_{0.5}$ on VLABench for long-horizon manipulation and generalization, and compact TurboVLA on RoboTwin 2.0 for bimanual manipulation under Clean and Randomized conditions.



\noindent \textbf{The CNN-based models.} Table~\ref{tab:CLIPORT_label} shows that both OptNet and DirNet improve the CNN-based CLIPORT's completion scores across diverse tasks and demonstration budgets. On \textit{align-rope} with 100 demonstrations, OptNet and DirNet increase the score from 62.4\% to 92.6\% and 93.6\%, corresponding to gains of 30.2 and 31.2 percentage points. The benefits also extend to limited-data settings and unseen object attributes. On \textit{packing-unseen-shapes} with only 10 demonstrations, OptNet and DirNet improve the score from 28.0\% to 38.0\% and 44.0\%, respectively. On the \textit{put-blocks-in-bowls-unseen-color} task with 1000 demonstrations, they raise the score from 18.5\% to 52.3\% and 52.2\%. These results suggest that the geometric canonicalization frame can facilitate spatial prediction and complement semantic representations when object attributes change, while preserving the base policy architecture.

\noindent \textbf{The Transformer-based models.} 

\begin{table*}[t]
\caption{Comparison of Transformer-based policies on (a) LIBERO, (b) VLABench, and (c) RoboTwin 2.0. We report success rates on LIBERO and RoboTwin 2.0, and intention score (IS) and progress score (PS) on VLABench. Avg. denotes the mean across the corresponding suites, tracks, or evaluation settings. Clean and Rand. denote the clean and randomized RoboTwin settings, respectively. Emb. PT. indicates additional embodied pretraining on robot data beyond RoboTwin 2.0. Bold and underlining indicate the best and second-best results in each column, respectively.}
\label{tab:transformer_comparison}
\label{tab:OpenVLA-OFT_label}
\label{tab:robotwin_comparison}
\label{tab:vlabench_comprehensive}
\centering
\fontsize{8}{8.5}\selectfont
\setlength{\tabcolsep}{1.2pt}
\renewcommand{\arraystretch}{1.0}
\newsavebox{\eqbotLiberoTableBox}
\newsavebox{\eqbotRobotwinTableBox}
\newsavebox{\eqbotVLABenchTopBox}
\newsavebox{\eqbotVLABenchBottomBox}
\sbox{\eqbotLiberoTableBox}{%
\begin{minipage}[t]{0.485\textwidth}
  \vspace{0pt}
  \centering
  \textbf{(a) LIBERO}\par\vspace{2pt}
  \begin{tabular*}{\linewidth}{@{\extracolsep{\fill}}lccccc@{}}
    \toprule
    Method & Spatial & Object & Goal & Long & Avg. \\
    \midrule
    SAVLA \cite{li2026savla} & 96.7 & 92.9 & 97.5 & 79.3 & 91.6 \\
    $\pi_0$ \cite{black2024pi_0} & 96.8 & \textbf{98.8} & 95.8 & 85.2 & 94.2 \\
    OpenVLA-OFT \cite{kim2025fine} & 96.6 & 97.8 & 97.6 & 94.1 & 96.5 \\
    \midrule
    Eq.Bot (OptNet) & \underline{97.9} & \underline{98.6} & \underline{98.4} & \underline{95.1} & \underline{97.5} \\
    Eq.Bot (DirNet) & \textbf{98.1} & \textbf{98.8} & \textbf{98.5} & \textbf{95.2} & \textbf{97.7} \\
    \bottomrule
  \end{tabular*}
\end{minipage}%
}
\sbox{\eqbotRobotwinTableBox}{%
\begin{minipage}[t]{0.485\textwidth}
  \vspace{0pt}
  \centering
  \setlength{\tabcolsep}{3pt}
  \renewcommand{\arraystretch}{1.05}
  \textbf{(c) RoboTwin 2.0}\par\vspace{2pt}
  \begin{tabular*}{\linewidth}{@{}l@{\hspace{12pt}}c@{\hspace{12pt}}c@{\hspace{18pt}}c@{\hspace{12pt}\extracolsep{\fill}}cc@{}}
    \toprule
    \multirow{3}{*}{Method} & \multirow{3}{*}{\shortstack{Emb.\\PT.}} & \multirow{3}{*}{\shortstack{Params\\(B) $\downarrow$}} & \multicolumn{3}{c}{Success Rate (\%) $\uparrow$} \\
    \cmidrule(lr){4-6}
    & & & Clean & Rand. & Avg. \\
    \midrule
    $\pi_{0.5}$ \cite{intelligence2025pi_} & \textcolor{green!60!black}{$\checkmark$} & 3.4 & 82.74 & 76.76 & 79.75 \\
    LingBot-VLA \cite{wu2026pragmatic} & \textcolor{green!60!black}{$\checkmark$} & 4.2 & 88.56 & 86.68 & 87.62 \\
    X-VLA \cite{zheng2026x} & \textcolor{green!60!black}{$\checkmark$} & 0.9 & 72.80 & 72.84 & 72.82 \\
    TurboVLA \cite{xie2026turbovla} & \textcolor{red}{$\times$} & \textbf{0.4} & 87.52 & 87.58 & 87.55 \\
    \midrule
    Eq.Bot (OptNet) & \textcolor{red}{$\times$} & \textbf{0.4} & \underline{89.60} & \underline{90.16} & \underline{89.88} \\
    Eq.Bot (DirNet) & \textcolor{red}{$\times$} & \underline{0.5} & \textbf{90.64} & \textbf{91.28} & \textbf{90.96} \\
    \bottomrule
  \end{tabular*}
\end{minipage}%
}
\sbox{\eqbotVLABenchTopBox}{%
\begin{minipage}[t]{0.485\textwidth}
  \vspace{0pt}
  \centering
  \textbf{(b) VLABench}\par\vspace{2pt}
  \begin{tabular*}{\linewidth}{@{\extracolsep{\fill}}lcccccc@{}}
    \toprule
    \multirow{2}{*}{Method} & \multicolumn{2}{c}{In-dist.} & \multicolumn{2}{c}{Category} & \multicolumn{2}{c}{Commonsense} \\
    \cmidrule(lr){2-3} \cmidrule(lr){4-5} \cmidrule(lr){6-7}
    & IS $\uparrow$ & PS $\uparrow$ & IS $\uparrow$ & PS $\uparrow$ & IS $\uparrow$ & PS $\uparrow$ \\
    \midrule
    $\pi_0$ \cite{black2024pi_0} & 67.8 & 62.7 & 44.0 & 33.6 & 54.9 & 43.0 \\
    ACoT-VLA \cite{zhong2026acot} & 79.8 & 66.1 & 54.1 & \textbf{38.9} & 52.3 & 37.8 \\
    $\pi_{0.5}$ \cite{intelligence2025pi_} & 73.2 & 59.6 & 50.0 & 32.9 & 61.5 & 43.3 \\
    \midrule
    Eq.Bot (OptNet) & \underline{81.8} & \underline{66.3} & \textbf{56.2} & \underline{37.4} & \underline{66.7} & \underline{44.1} \\
    Eq.Bot (DirNet) & \textbf{82.6} & \textbf{67.1} & \underline{55.8} & 36.2 & \textbf{67.6} & \textbf{45.9} \\
    \bottomrule
  \end{tabular*}
\end{minipage}%
}
\sbox{\eqbotVLABenchBottomBox}{%
\begin{minipage}[t]{0.485\textwidth}
  \vspace{0pt}
  \centering
  \begin{tabular*}{\linewidth}{@{\extracolsep{\fill}}lcccccc@{}}
    \toprule
    \multirow{2}{*}{Method} & \multicolumn{2}{c}{Instruction} & \multicolumn{2}{c}{Texture} & \multicolumn{2}{c}{Avg.} \\
    \cmidrule(lr){2-3} \cmidrule(lr){4-5} \cmidrule(lr){6-7}
    & IS $\uparrow$ & PS $\uparrow$ & IS $\uparrow$ & PS $\uparrow$ & IS $\uparrow$ & PS $\uparrow$ \\
    \midrule
    $\pi_0$ \cite{black2024pi_0} & 58.0 & \underline{38.7} & 50.6 & 42.5 & 55.1 & 44.1 \\
    ACoT-VLA \cite{zhong2026acot} & 56.8 & \textbf{39.6} & \textbf{74.6} & \textbf{54.6} & 63.5 & \textbf{47.4} \\
    $\pi_{0.5}$ \cite{intelligence2025pi_} & 60.2 & 27.2 & 60.1 & 48.9 & 61.0 & 42.4 \\
    \midrule
    Eq.Bot (OptNet) & \underline{65.8} & 30.3 & 61.4 & 49.2 & \underline{66.4} & 45.5 \\
    Eq.Bot (DirNet) & \textbf{67.6} & 31.8 & \underline{61.9} & \underline{49.4} & \textbf{67.1} & \underline{46.1} \\
    \bottomrule
  \end{tabular*}
\end{minipage}%
}
\newlength{\eqbotTableHeight}
\newlength{\eqbotRightTableHeight}
\setlength{\eqbotTableHeight}{\dimexpr
  \ht\eqbotLiberoTableBox+\dp\eqbotLiberoTableBox+
  \ht\eqbotRobotwinTableBox+\dp\eqbotRobotwinTableBox+0.5em\relax}
\setlength{\eqbotRightTableHeight}{\dimexpr
  \ht\eqbotVLABenchTopBox+\dp\eqbotVLABenchTopBox+
  \ht\eqbotVLABenchBottomBox+\dp\eqbotVLABenchBottomBox+0.5em\relax}
\ifdim\eqbotRightTableHeight>\eqbotTableHeight
  \setlength{\eqbotTableHeight}{\eqbotRightTableHeight}
\fi
\noindent\makebox[\textwidth][c]{%
  \vbox to\eqbotTableHeight{%
    \offinterlineskip
    \copy\eqbotLiberoTableBox
    \vfill
    \copy\eqbotRobotwinTableBox
    \vskip0pt
  }\hfill
  \vbox to\eqbotTableHeight{%
    \offinterlineskip
    \copy\eqbotVLABenchTopBox
    \vfill
    \copy\eqbotVLABenchBottomBox
    \vskip0pt
  }%
}
\vspace{-4ex}
\end{table*}

\noindent \textit{\textbf{LIBERO.}} Table~\ref{tab:OpenVLA-OFT_label}(a) shows both Eq.Bot variants improve OpenVLA-OFT across all four suites. Average success increases from 96.5\% to 97.5\% with OptNet and 97.7\% with DirNet. The largest gains occur on LIBERO-Spatial, where success increases from 96.6\% to 97.9\% and 98.1\%, respectively, consistent with canonicalization providing a geometric reference for action prediction. Both variants also exceed the 91.6\% average success rate reported by SAVLA \cite{li2026savla}, an end-to-end symmetry-aware VLA with a specialized equivariant action head. These gains support coupling observation canonicalization with action recovery for direct regression while preserving the base policy architecture.

\noindent \textit{\textbf{VLABench.}} Both variants improve the flow matching based $\pi_{0.5}$ performance across all five tracks (Table~\ref{tab:vlabench_comprehensive}(b)), with substantial gains on \textit{In-dist.} and \textit{Instruction}. On \textit{In-dist.}, OptNet improves IS/PS by 8.6/6.7 percentage points and DirNet by 9.4/7.5 points; on \textit{Instruction}, the corresponding gains are 5.6/3.1 and 7.4/4.6 points. Improvements in both target-directed behavior and task progress are consistent with the benefit of the canonicalization frame. Both variants also exceed the recently reported ACoT-VLA \cite{zhong2026acot} results in IS and PS of \textit{In-dist.} and IS of \textit{Instruction}, with DirNet improving the latter by 10.8 points. However, lower \textit{Instruction} PS and \textit{Texture} scores indicate remaining challenges in task execution and appearance generalization.

\noindent \textit{\textbf{RoboTwin 2.0.}} 
Eq.Bot also improves compact TurboVLA across 50 bimanual tasks (Table~\ref{tab:robotwin_comparison}(c)). Average success increases from 87.55\% to 89.88\% with OptNet and 90.96\% with DirNet. Gains under Randomized conditions reach 2.58 and 3.70 percentage points, exceeding the respective Clean gains of 2.08 and 3.12 points, supporting the usefulness of canonicalization under environmental variation. Both variants also exceed LingBot-VLA's reported average success of 87.62\%, with substantially smaller parameter counts (0.4B and 0.5B versus 4.2B) and without additional embodied pretraining beyond RoboTwin 2.0. These results extend the canonicalization to compact bimanual policies while preserving the base policy architecture.

\subsection{Ablation Study}

\begin{table}[t]
\centering
\setlength{\tabcolsep}{4pt}
\caption{Ablations on Ravens. Completion scores (\%) with 10 demos per task are reported; bold and underlining indicate the best and second-best rotation-group results, respectively.}
\resizebox{\linewidth}{!}{%
  \begin{tabular}{lcccccccc}
    \toprule
    & \multicolumn{4}{c}{pack \textcolor{blue}{seen} objects} & \multicolumn{4}{c}{assemble \textcolor{blue}{seen} kits} \\
    \cmidrule(lr){2-5} \cmidrule(lr){6-9}
    Method & $C_2$ & $C_4$ & $C_8$ & $C_n$ & $C_2$ & $C_4$ & $C_8$ & $C_n$ \\
    \midrule
    Eq.Bot (OptNet-CNN) & 53.7 & \underline{56.0} & 55.5 & \textbf{56.7} & 15.2 & \textbf{17.8} & 16.8 & \underline{17.6} \\
    Eq.Bot (OptNet-Res50) & 54.1 & \textbf{57.5} & \underline{57.0} & 56.3 & 16.4 & \textbf{18.3} & 18.0 & \underline{18.1} \\
    Eq.Bot (DirNet) & 54.6 & \textbf{58.2} & 57.6 & \underline{57.9} & 17.6 & \textbf{19.4} & 18.7 & \underline{19.0} \\
    \midrule
    & \multicolumn{4}{c}{pack \textcolor{red}{unseen} objects} & \multicolumn{4}{c}{assemble \textcolor{red}{unseen} kits} \\
    \cmidrule(lr){2-5} \cmidrule(lr){6-9}
    Method & $C_2$ & $C_4$ & $C_8$ & $C_n$ & $C_2$ & $C_4$ & $C_8$ & $C_n$ \\
    \midrule
    Eq.Bot (OptNet-CNN) & 44.5 & \underline{45.3} & 43.9 & \textbf{46.6} & 17.8 & \textbf{19.6} & 18.4 & \underline{19.1} \\
    Eq.Bot (OptNet-Res50) & 44.9 & \textbf{45.9} & 44.6 & \underline{45.2} & 18.3 & \textbf{21.6} & 19.6 & \underline{21.4} \\
    Eq.Bot (DirNet) & 45.9 & \textbf{47.3} & \underline{46.5} & 45.7 & 18.9 & \textbf{22.4} & \underline{20.7} & 18.8 \\
    \bottomrule
  \end{tabular}}
\label{tab:ablation_study}
\vspace{-2ex}
\end{table}

\begin{figure*}[t]
    \centering
    \includegraphics[width=1\linewidth]{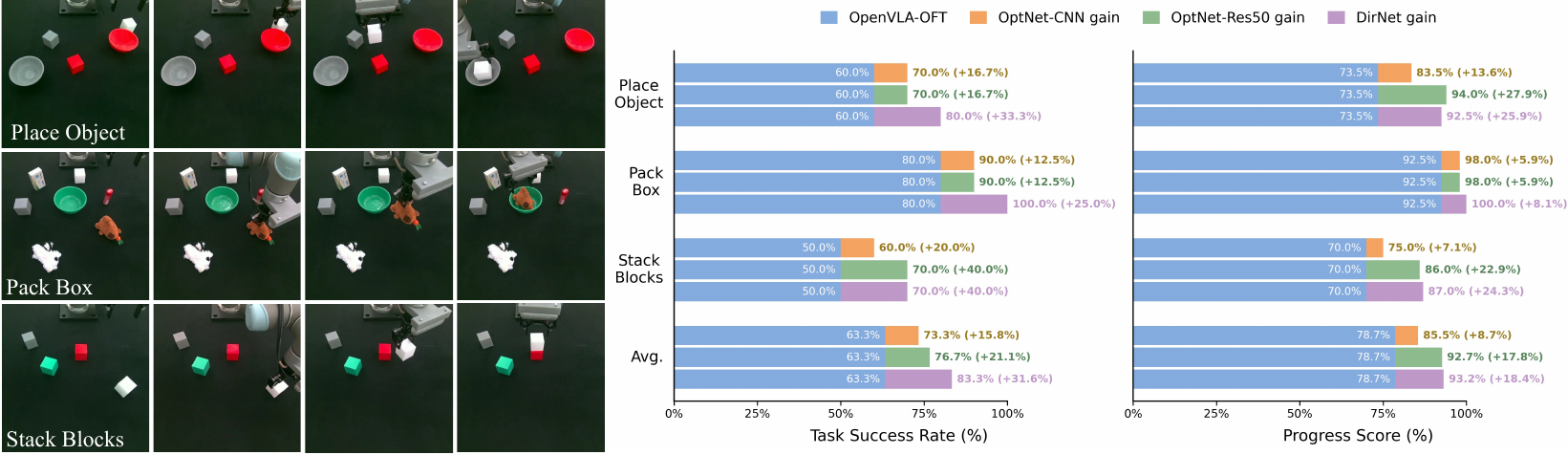}
    \caption{Real-world results. Left: visualization of the three manipulation tasks. Center: task success rates. Right: progress scores.}
    \label{fig:real_full}
\vspace{-2ex}
\end{figure*}

\noindent \textbf{Canonicalization Network Analysis.} To evaluate different canonicalizer architectures, we additionally implement OptNet with a lightweight conventional CNN (OptNet-CNN), alongside OptNet-Res50 and DirNet. As shown in Table~\ref{tab:ablation_study}, completion scores on \textit{assemble-seen-kits} with $C_4$ increase from 17.8\% with OptNet-CNN to 18.3\% with OptNet-Res50 and 19.4\% with DirNet. This suggests that both encoder expressiveness and explicit geometric inductive bias can benefit canonicalization. DirNet also improves over OptNet-Res50 on LIBERO (97.7\% versus 97.5\%), extending this comparison to the Transformer-based policy. Real-world results show that even the lightweight OptNet-CNN improves average success from 63.3\% to 73.3\% and average progress from 78.7\% to 85.5\% (Figure~\ref{fig:real_full}). On \textit{Stack Blocks}, which requires precise alignment, OptNet-Res50 and DirNet achieve further improvements, reaching progress scores of 86.0\% and 87.0\%, compared with 75.0\% for OptNet-CNN. Together, these findings support the effectiveness of canonicalization across different network designs, with richer representations and explicit equivariance offering additional benefits for precise manipulation.

\noindent \textbf{Rotation Group Analysis.} A critical design choice in our framework is the rotation group used for canonicalization, which determines the coverage and granularity of orientation alignment. As shown in Table \ref{tab:ablation_study}, we conduct a comprehensive ablation study across $C_2$, $C_4$, $C_8$, and $C_n$ to investigate the trade-off between geometric precision and estimation robustness.

The results show that $C_4$ provides the most consistent performance across the tested settings. A plausible explanation is that it combines useful rotational coverage with preservation of fine spatial detail. Unlike $C_2$, $C_4$ includes quarter-turn rotations and can align observations differing by $90^\circ$. It is also the largest planar rotation group that preserves the square pixel lattice \cite{cohen2016group}, allowing transformations without angular interpolation. The additional angles in $C_8$ and $C_n$ generally require resampling of images and affordance maps, which may perturb object boundaries and placement cues critical to packing and assembly. Finer orientation distinctions may also be harder to learn with limited demonstrations. For instance, on \textit{assemble-unseen-kits} with 10 demonstrations, Eq.Bot (DirNet) achieves 22.4\% with $C_4$, compared with 20.7\% for $C_8$ and 18.8\% for $C_n$. Thus, finer rotational coverage does not consistently improve performance, supporting $C_4$ as a practical balance between coverage, geometric fidelity, and learnability.

\subsection{Real-World Experiments}

We evaluate Eq.Bot on three UR5e tasks with complementary geometric requirements: \textit{Place Object} tests understanding of relative spatial relationships, \textit{Pack Box} evaluates target selection among multiple objects, and \textit{Stack Blocks} assesses precise placement and alignment. All tasks vary object positions and require avoiding collisions with non-target objects, with \textit{Stack Blocks} demanding the highest alignment precision. OpenVLA-OFT and its OptNet-CNN, OptNet-Res50, and DirNet variants use the same datasets and are evaluated over 10 trials per task. We report collision-free task success and progress scores, while more evaluation details are provided in the appendix.

Figure~\ref{fig:real_full} shows improvements in both metrics on every task for all three variants. DirNet achieves the highest average success rate, increasing it from 63.3\% to 83.3\%, with the average progress score also increasing from 78.7\% to 93.2\%. OptNet-CNN and OptNet-Res50 also improve average success to 73.3\% and 76.7\%, respectively. This indicates that the benefits extend to both optimization-based and direct canonicalization. The results on \textit{Stack Blocks} highlight the value of canonicalization for precise manipulation. OptNet-Res50 and DirNet both raise success from 50.0\% to 70.0\%, while increasing progress from 70.0\% to 86.0\% and 87.0\%, respectively. Thus, canonicalization improves both completion of the final objective and progress through intermediate stages, supporting its practical value for physical manipulation without modifying the base policy.

\subsection{Inference Efficiency}

\begin{table}[t] 
 \centering 
 \caption{Latency and overhead of CLIPORT and OpenVLA-OFT with different canonicalization networks.}
 \resizebox{1\columnwidth}{!}{%
   \begin{tabular}{lcccc} 
     \toprule 
     \multirow{2}{*}{Method} & \multicolumn{2}{c}{CLIPORT} & \multicolumn{2}{c}{OpenVLA-OFT} \\ 
     \cmidrule(lr){2-3} \cmidrule(lr){4-5} 
     & Latency (ms) & Overhead & Latency (ms) & Overhead \\ 
     \midrule 
     Base Policy & 81.99 & - & 110.81 & - \\ 
     \midrule 
     Eq.Bot (OptNet-CNN) & 87.77 & +7.05\% & 113.24 & +2.19\% \\ 
     Eq.Bot (OptNet-Res50) & 91.26 & +11.31\% & 116.12 & +4.79\% \\ 
     Eq.Bot (DirNet) & 178.36 & +117.54\% & 135.66 & +22.43\% \\ 
     \bottomrule 
   \end{tabular}} 
 \label{tab:inference_time} 
 \vspace{-2.5ex}
 \end{table}

Table \ref{tab:inference_time} reports inference latency on a single NVIDIA V100 GPU, with overhead measured relative to the corresponding base policy. OptNet-CNN and OptNet-Res50 add 5.78--9.27 ms to CLIPORT inference (7.05--11.31\% overhead) and 2.43--5.31 ms to OpenVLA-OFT (2.19--4.79\%). DirNet incurs higher overhead: 117.54\% for CLIPORT and 22.43\% for OpenVLA-OFT. Thus, the cost of canonicalization depends on both the canonicalizer and the base policy. OptNet-CNN introduces the least overhead, while OptNet-Res50 combines competitive task performance with substantially lower latency than DirNet. These measurements show that optimization-based canonicalization can retain much of the observed performance benefit with modest overhead, supporting selection between task performance and inference efficiency.

 \vspace{-1ex}
\section{Conclusion}

In this work, we introduce Eq.Bot, a model-agnostic canonicalization framework designed to bridge the geometric gap in existing robotic manipulation policies. By coupling observation alignment with representation-specific action recovery through a shared estimated frame, Eq.Bot incorporates geometric equivariance while preserving the underlying policy architecture. DirNet and OptNet provide alternative implementations that learn canonical observation through equivariant features and conventional encoder features, respectively. Experiments across four simulation benchmarks and three real-world tasks demonstrate improved manipulation performance across CNN- and Transformer-based policies of different scales, including compact bimanual models. OptNet retains competitive task performance with lower inference overhead than DirNet, offering a practical performance-latency trade-off. Together, these findings support learned canonicalization as an effective way to connect existing policy representations with the geometric requirements of manipulation.

\section*{LLM Usage Statement}

The authors used OpenAI Codex to assist with manuscript drafting and editing, literature searches and related-work organization, interpretation and presentation of experimental results, and preparation of visualization scripts and figure layouts. OpenAI's image-generation tool was used to create and refine the conceptual human-profile illustration. The authors take full responsibility for all submitted content.

{
    \small
    \bibliographystyle{ieeenat_fullname}
    \bibliography{main}
}

\clearpage
\setcounter{page}{1}

\section*{A. Equivariance Analysis of Existing VLA Model Architectures} \label{sec:Equivariance_Analysis}

We examine why standard CNN- and Transformer-based policies fail to enforce rotation equivariance by construction. Following the main text, we hold the language instruction fixed and consider a rotation $g\in G$, with $G=C_4$ in the main experiments. An equivariant policy must satisfy $\pi(g\cdot I,\ell)=\rho_{\mathcal{A}}(g)\pi(I,\ell)$.

\subsection*{A.1 CNN-based Policies}

We commence by analyzing the intrinsic geometric limitations of standard Convolutional Neural Networks (CNNs), subsequently extending our analysis to CNN-based VLA Policies such as CLIPort \cite{shridhar2022cliport}.

\noindent \textbf{Convolution and Downsampling.}
Standard convolutions are translation-equivariant under suitable boundary conditions \cite{cohen2016group}. For rotations, let $\mathcal{T}_g$ denote geometric rotation, $(\mathcal{T}_g I)(x)=I(g^{-1}x)$, and let $C_W$ denote the usual discrete cross-correlation with kernel $W$. For a rotation that preserves the pixel lattice and with compatible boundary handling, changing variables in the correlation sum gives:
\begin{equation}
    C_W(\mathcal{T}_g I)=\mathcal{T}_g C_{\mathcal{T}_{g^{-1}}W}(I).
\end{equation}
Thus, for output channels that transform only by geometric rotation, equivariance requires the kernel to be invariant. Unconstrained kernels do not satisfy this condition, even for $C_4$. Strided convolutions and pooling can also introduce aliasing and reduce consistency under transformations that do not preserve the sampling grid \cite{zhang2019making}.

\noindent \textbf{Feature Fusion in CLIPORT.}
CLIPORT combines a CLIP-based semantic stream with a convolutional spatial stream in the Transporter policy \cite{shridhar2022cliport,radford2021learning,zeng2021transporter}. The semantic stream processes RGB observations and uses the language embedding to condition dense visual features. The pick branch adds the spatial and semantic maps, $F_{\mathrm{pick}}(I,\ell)=F_{\mathrm{spat}}(I)+F_{\mathrm{sem}}(I,\ell)$. Since $\mathcal{T}_g$ acts linearly on these maps, the equivariance residual decomposes as:
\begin{equation}
\begin{aligned}
    &F_{\mathrm{pick}}(g\cdot I,\ell)-\mathcal{T}_gF_{\mathrm{pick}}(I,\ell)\\
    &= \big[F_{\mathrm{spat}}(g\cdot I)-\mathcal{T}_gF_{\mathrm{spat}}(I)\big]\\
    &\quad + \big[F_{\mathrm{sem}}(g\cdot I,\ell)-\mathcal{T}_gF_{\mathrm{sem}}(I,\ell)\big].
\end{aligned}
\end{equation}
If the spatial branch is rotation-equivariant, the first term vanishes. The broadcast language embedding is constant for a fixed instruction, so pointwise language gating commutes with spatial permutation. The remaining condition is that the dense semantic features must follow the same spatial transformation law, while CLIP pretraining and standard fusion layers do not impose.

\subsection*{A.2 Transformer-based Policies}

We analyze how patch embeddings and absolute positional encodings in standard Vision Transformers respond to image rotations, and what this implies for rotation equivariance in VLA policies such as OpenVLA-OFT and $\pi_{0.5}$.

\noindent \textbf{Patch Embeddings and Positional Encodings.}
Standard Vision Transformers embed image patches and add absolute positional embeddings \cite{dosovitskiy2020image}. Rotations that do not preserve the patch grid mix content across patch boundaries, so reindexing patches does not in general recover the transformed input. Even when a rotation preserves the grid, it induces both a permutation $\sigma_g$ of patch indices and a permutation $P_g$ of pixels within each patch. For a linear patch projection $E$, patch vector $x_i$, and positional embedding $e_i$, the corresponding input tokens are:
\begin{equation}
    z_i=E(x_i)+e_i,\qquad z'_{\sigma_g(i)}=E(P_gx_i)+e_{\sigma_g(i)}.
\end{equation}
For the token transformation $z'_{\sigma_g(i)}=Q_gz_i$ to hold for all patch contents, the projection must satisfy $EP_g=Q_gE$ and the positional embeddings must satisfy $e_{\sigma_g(i)}=Q_ge_i$, where $Q_g$ acts on token channels. However, sharing a patch projection across locations and learning absolute positions do not enforce these conditions. Self-attention can be equivariant to permutations of its input tokens, but rotating the image need not produce the corresponding permutation of those tokens.

Eq.Bot uses a shared estimated frame to align observations before policy inference and map predictions back according to action representation. As shown in Section~\ref{method}, consistent frame estimation and compatible observation-action mappings yield equivariance of the composite policy while preserving the base architecture.

\section*{B. Action Representations and Recovery}

Eq.Bot uses the same frame estimate for observation canonicalization and action recovery. Let $g=g_{\mathrm{est}}$ map the canonical frame to the original frame and let $\mathsf D_g$ denote the corresponding action map in Eq.~\ref{eq:action_mapping}. For each native action representation, the same convention maps training targets into canonical coordinates via $\mathsf D_{g^{-1}}$ and predictions back via $\mathsf D_g$. Table~\ref{tab:action_recovery_interfaces} summarizes the maps, which preserve the base policy's architecture and output representation.

\begin{table*}[t]
    \centering
    \small
    \setlength{\tabcolsep}{4pt}
    \renewcommand{\arraystretch}{1.15}
    \caption{Native action representations and recovery operators for the evaluated policies. Each operator maps predictions from the canonical frame to the original coordinates while preserving gripper commands.}
    \label{tab:action_recovery_interfaces}
    \begin{tabular}{@{}p{0.21\textwidth}p{0.35\textwidth}p{0.40\textwidth}@{}}
        \toprule
        Policy / setting & Action representation & Recovery in the original coordinates \\
        \midrule
        CLIPORT / Ravens & Pick/place score maps or pixel coordinates & Rotate maps, points, and absolute headings. \\
        \midrule
        OpenVLA-OFT / LIBERO & World-frame translation and rotation increments & Rotate both increments by $R_g$. \\
        \midrule
        OpenVLA-OFT / real world & Absolute TCP pose in the robot base frame & Compose $T=H_gT_c$ and re-encode the rotation vector. \\
        \midrule
        $\pi_{0.5}$ / VLABench & Absolute end-effector poses in world-frame & Compose $T=H_gT_c$ and re-encode the Euler angles. \\
        \midrule
        TurboVLA / RoboTwin 2.0 & Bimanual joint-position targets & Apply $H_g$ via per-arm FK and IK. \\
        \bottomrule
    \end{tabular}
\end{table*}

\noindent \textbf{Shared Coordinate Convention.}
A rotation by $\alpha$ about a workspace center $c$ has the homogeneous form:
\begin{equation}
    H_g=
    \begin{bmatrix}
        R_g & t_g\\
        0 & 1
    \end{bmatrix},
    \qquad R_g=R_z(\alpha),\quad t_g=c-R_gc.
    \label{eq:recovery_workspace_frame}
\end{equation}
If $T_{WF}$ maps camera or robot base frame $F$ to workspace $W$, the equivalent transform in $F$ is $T_{WF}^{-1}H_gT_{WF}$. Observation warps must represent this shared workspace transform across views. Planar homographies are exact only on the assumed plane, and cropping, occlusion, or resampling can all further affect observation alignment.

\noindent \textbf{Pixel Coordinates and Score Maps.}
For CLIPORT, pixel vectors use $(\mathrm{column},\mathrm{row})$, with right and down positive. Let $\bar R_g$ be the canonical-to-original rotation in this pixel basis, $c_{\mathrm{pix}}=(S/2,S/2)$ the rotation center on the square padded grid of side $S$, and $b$ the padding offset. A canonical pixel $\mathbf u_c$ maps to the original grid as:
\begin{equation}
    \mathbf u=c_{\mathrm{pix}}+\bar R_g(\mathbf u_c-c_{\mathrm{pix}})-b.
    \label{eq:recovery_pixels}
\end{equation}
Spatial scores follow the same coordinate map, with each original pixel assigned the score at its canonical preimage. Absolute headings rotate by the angle of $\bar R_g$. The relative placement angle $\beta_{\mathrm{place}}-\beta_{\mathrm{pick}}$ stays fixed modulo $2\pi$ because the same rotation is applied to both headings. Quarter turns act by exact permutations on a pixel grid closed under $C_4$, while cropping and resampling can limit invertibility.

\noindent \textbf{Cartesian Poses and Increments.}
In our experiments, OpenVLA-OFT uses absolute tool center point (TCP) poses in the robot base frame for real-world tasks and world-frame translation and rotation increments for LIBERO. For an absolute end-effector pose $T_c$ with canonical position $p_c$ and orientation $R_c$, we express $H_g$ in the same reference frame as $T_c$ and recover the pose as:
\begin{equation}
    \begin{aligned}
        T&=H_gT_c,\qquad T_c=H_g^{-1}T,\\
        p&=R_gp_c+t_g,\qquad R=R_gR_c.
    \end{aligned}
    \label{eq:recovery_absolute_pose}
\end{equation}
The recovered orientation $R$ is expressed in the policy's native parameterization, such as a rotation vector or extrinsic XYZ Euler angles. For world-frame increments, let $\Delta R_c=R_{\mathrm{tar},c}R_{\mathrm{cur},c}^{\top}$, where $R_{\mathrm{tar},c}$ and $R_{\mathrm{cur},c}$ are the canonical target and current orientations. The common positional offset cancels between the current and target positions, and $R_g$ conjugates the relative rotation:
\begin{equation}
    \Delta p=R_g\Delta p_c,\qquad
    \Delta R=R_g\Delta R_cR_g^{\top}.
    \label{eq:recovery_delta_pose}
\end{equation}
The corresponding rotation-vector increment is $\Delta\omega=R_g\Delta\omega_c$. Before recovery, we undo any action normalization or controller scaling to express positions and displacements in meters and rotation parameters in radians.

\noindent \textbf{Compatibility with Flow Matching.}
For $\pi_{0.5}$, let $N$ denote the invertible action normalization and $A_{\mathrm{gt}}$ the demonstrated action chunk. Canonicalization precedes normalization in the flow-matching objective:
\begin{equation}
    \begin{aligned}
        Z_0 &= N\!\left(\mathsf D_{g^{-1}}(A_{\mathrm{gt}})\right),\\
        Z_t &= (1-t)Z_0+t\epsilon,\qquad U_t=\epsilon-Z_0,\\
        \mathcal L_{\mathrm{FM}}
        &=\mathbb E\!\left[\left\|v_{\theta}(Z_t,t\mid\mathcal O_c,\ell)-U_t\right\|^2\right].
    \end{aligned}
    \label{eq:recovery_flow_matching}
\end{equation}
Here, $\epsilon$ is Gaussian noise and $\mathcal O_c$ contains canonical observations and robot states. Following the openpi convention, sampling integrates from $t=1$ to $t=0$ and recovers the chunk as $\mathsf D_g(N^{-1}(\widehat Z_0))$. The geometric mapping acts only on VLABench's seven native action dimensions, excluding model padding. A pathwise comparison uses shared canonical noise for paired transformed inputs.

\noindent \textbf{Joint-Space Recovery.}
For TurboVLA, the end-effector pose of arm $j$ in the shared workspace is $K_j(q)=T_{WB_j}\operatorname{FK}_j(q)$, where $T_{WB_j}$ maps it from the arm's base frame to the workspace frame. If $K_j$ is one-to-one on a selected joint domain $\mathcal U_j$ and $K_j(\mathcal U_j)$ is invariant under the considered transforms $H_g$, the recovery operator is:
\begin{equation}
    \mathsf D_g^{(j)}(q)=K_j^{-1}\!\left(H_gK_j(q)\right).
    \label{eq:recovery_joint_space}
\end{equation}
The two arms use the same $H_g$, with inverse kinematics evaluated on the selected branch of each arm. On these domains, composition of workspace transforms gives:
\begin{equation}
    \mathsf D_g^{(j)}\circ\mathsf D_h^{(j)}=\mathsf D_{gh}^{(j)},
    \qquad
    \mathsf D_{g^{-1}}^{(j)}\circ\mathsf D_g^{(j)}=\mathrm{id}.
    \label{eq:recovery_joint_group_law}
\end{equation}
The corresponding inverse maps canonicalize joint targets and state inputs. For each mapped configuration, the implementation checks domain membership, $K_j^{-1}(K_j(q))=q$, and reconstruction of the transformed pose, using a default numerical tolerance of $10^{-5}$. Reachability, joint limits, and singularities restrict the valid domain.

\section*{C. Experiment Details} \label{appendix:experiment_details}

This section describes the datasets, implementation, and evaluation protocols for Eq.Bot. Supplementary results further assess how its shared interface for observation canonicalization and action recovery performs across manipulation policies and tasks.

\subsection*{C.1 Datasets Description}

\noindent \textbf{Ravens Benchmark.}
We use the language-conditioned Ravens demonstrations released with CLIPort \cite{shridhar2022cliport}. The 18 evaluated task settings cover object packing, kit assembly, block stacking, etc. Demonstrations contain RGB-D observations, language goals, and pick-and-place poses at manipulation-primitive boundaries. We use the benchmark's training and evaluation splits, including seen and unseen color or object-category settings, with training budgets of 1, 10, 100, and 1000 demonstrations per task.

\noindent \textbf{LIBERO Benchmark.}
We use the modified, no-op-filtered LIBERO datasets released in RLDS format with OpenVLA \cite{kim2024openvla}, covering \textit{LIBERO-Spatial}, \textit{LIBERO-Object}, \textit{LIBERO-Goal}, and \textit{LIBERO-Long}, with 10 tasks per suite \cite{liu2023libero}. These suites emphasize spatial relationships, target objects, task goals, and longer action sequences, respectively. Each trajectory provides external and wrist-view RGB observations, robot states, language instructions, and 7-dimensional actions comprising incremental end-effector motion and a gripper command. Training uses the corresponding suite's demonstrations, while evaluation uses its benchmark-provided initial states.

\noindent \textbf{VLABench Benchmark.}
We use the official VLABench primitive task fine-tuning dataset \cite{zhang2025vlabench}, containing 10 tasks with 500 expert demonstrations per task. Following the VLABench-openpi data interface, raw HDF5 trajectories are converted to LeRobot format, retaining RGB views, robot states, continuous action targets, and language instructions. Evaluation follows five released track configurations: \textit{In-distribution}, \textit{Category}, \textit{Commonsense}, \textit{Instruction}, and \textit{Texture}. These assess in-domain task learning, object-category and instance generalization, commonsense grounding, semantic instruction understanding, and robustness to changes in backgrounds and tabletop textures, respectively. The same training data and track configurations are used for the base policy and its Eq.Bot variants.

\noindent \textbf{RoboTwin 2.0 Benchmark.}
We use both the Clean and Randomized training datasets for all 50 RoboTwin 2.0 tasks \cite{chen2025robotwin,xie2026turbovla}: 50 clean and 500 randomized demonstrations per task, totaling 2,500 and 25,000 trajectories, respectively. The LeRobot-formatted data provide a head-camera view, two wrist-camera views, task instructions, 14-dimensional bimanual states, and absolute joint-position and gripper targets. These records support training with both nominal and randomized scene conditions.

\noindent \textbf{Real-World Dataset.}
Our dataset contains 50 teleoperated expert demonstrations for each of three tasks, totaling 150 trajectories: \textit{Place Object} (move the white block next to the red bowl into the gray bowl), \textit{Pack Box} (place the brown toy in the green basin), and \textit{Stack Blocks} (place the white block on the red block). Each trajectory records external and wrist-view RGB observations, the task instruction, end-effector poses, and gripper states. We convert the recordings to RLDS format and construct 7-dimensional action labels from the recorded state trajectories, encoding absolute end-effector pose targets and gripper commands.

\subsection*{C.2 Implementation Details}

\noindent \textbf{CLIPORT on Ravens.}
We train a separate policy for each Ravens task setting and demos for 200,000 iterations on a single NVIDIA V100 GPU (32 GB), using a batch size of 1. RGB-D observations are fused into $320\times160\times6$ top-down representations. The pretrained CLIP visual and text encoders remain frozen, while the remaining policy parameters are optimized using Adam at a learning rate of $10^{-4}$. For our canonicalization modules, a higher learning rate of $1 \times 10^{-3}$ was employed to promote faster convergence of the canonicalization network, ensuring the policy receives stabilized inputs early in the training process.

\noindent \textbf{OpenVLA-OFT on LIBERO.}
We initialize from the pretrained OpenVLA-7B base checkpoint and fine-tune independently on each LIBERO suite \cite{kim2025fine}. LoRA is applied to linear layers with rank 32, alpha of 16, and zero dropout; the original backbone weights remain frozen. We jointly optimize the adapters, action head, proprioceptive projector, and the canonicalization module for Eq.Bot. All models are fine-tuned for 150,000 steps on two NVIDIA A100 GPUs (40 GB each), with a per-GPU batch size of 2 and an effective batch size of 4. We use AdamW with an initial learning rate of $5\times10^{-4}$ for all trainable modules, reduced to $5\times10^{-5}$ after 100,000 steps.

\noindent \textbf{$\pi_{0.5}$ on VLABench.}
We fine-tune the released $\pi_{0.5}$ base checkpoint \cite{intelligence2025pi_} on VLABench using the \texttt{openpi} implementation. Training runs for 60,000 optimizer steps on two NVIDIA A100 GPUs (40 GB each), with a global batch size of 64. We apply LoRA to the attention and feed-forward layers of the 2B language backbone and 300M action expert, with $(r,\alpha)=(16,16)$ and $(32,32)$, respectively, while freezing the non-adapter parameters within these Transformer components. Images are processed at $224\times224$, the action horizon is set to 10, and discrete state tokenization is disabled. We use AdamW with $(\beta_1,\beta_2)=(0.9,0.95)$, $\epsilon=10^{-8}$, weight decay $10^{-10}$, and gradient-norm clipping at 1.0. The learning rate increases to $5\times10^{-5}$ over 1,000 warm-up steps and remains constant thereafter.

\noindent \textbf{TurboVLA on RoboTwin 2.0.}
We use the DINOv3 ViT-L visual encoder, frozen BERT instruction encoder, and ACT action head of TurboVLA \cite{xie2026turbovla}. The policy receives three RGB views at $224\times224$ and a 14-dimensional robot state, predicting 50-step chunks of 14-dimensional absolute joint position and gripper actions. Training covers all 50 tasks with uniform task sampling and a Clean:Randomized sampling ratio of $1{:}10$. We train for 150,000 optimizer steps on two NVIDIA GeForce RTX 2080 Ti GPUs (11 GB each), using FP32 and DeepSpeed ZeRO-2. Each GPU runs one process with a batch size of 4 and one gradient-accumulation step. Trainable policy modules are optimized with an L1 behavior-cloning loss and AdamW with $(\beta_1,\beta_2)=(0.9,0.95)$, $\epsilon=10^{-8}$, and weight decay $10^{-10}$. The learning rate increases linearly to $5\times10^{-5}$ over 1,000 warm-up steps and then stays constant. Gradients are clipped at norm 1.0, and EMA weights are maintained on the CPU with decay 0.999.

\begin{figure}[t]
    \centering
    \includegraphics[width=\linewidth]{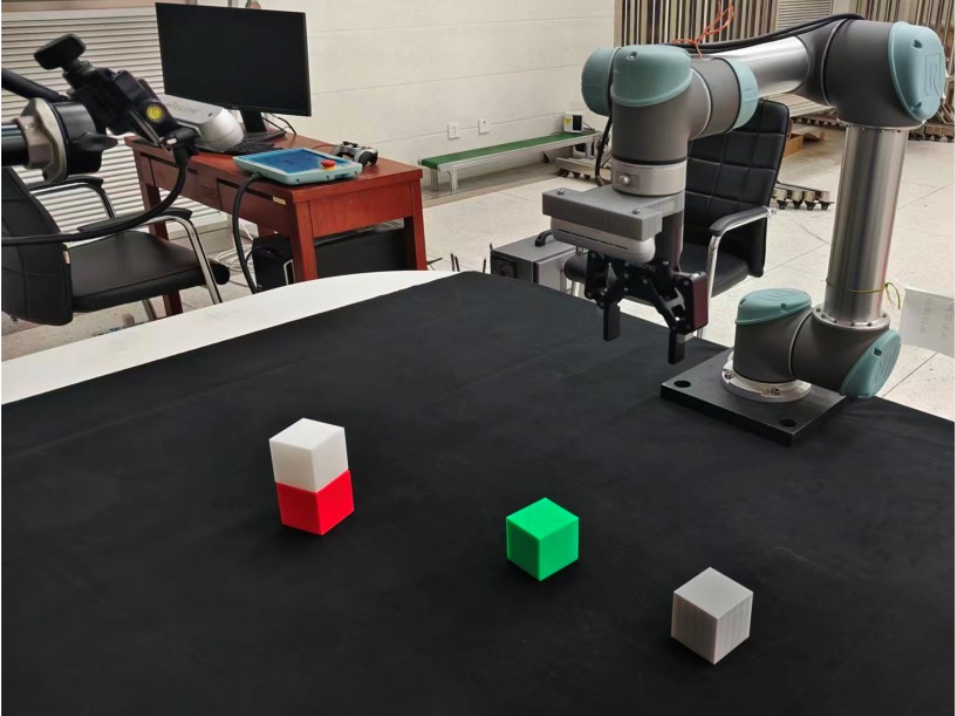}
    \caption{Real-world deployment setup.}
    \label{fig:real_world_setup}
    \vspace{-2ex}
\end{figure}

\noindent \textbf{Real-World Experimental Setup.}
As shown in Figure~\ref{fig:real_world_setup}, physical experiments use a UR5e robotic arm equipped with a Robotiq 2F-85 parallel gripper. Two Intel RealSense cameras provide fixed external and wrist-mounted RGB views. Through teleoperation, we collect 50 expert demonstrations for each of \textit{Place Object}, \textit{Pack Box}, and \textit{Stack Blocks}. A ROS2-based pipeline records paired camera views, task instructions, end-effector poses, and gripper states. OpenVLA-OFT and its Eq.Bot variants use the same demonstration sets and follow the OpenVLA-OFT fine-tuning settings described above. Deployment runs on a server equipped with two NVIDIA GeForce RTX 2080 Ti GPUs, each with 11 GB of memory.

\subsection*{C.3 Evaluation Protocols}

\begin{table*}[t]
    \centering
    \small
    \setlength{\tabcolsep}{6pt}
    \renewcommand{\arraystretch}{1.15}
    \caption{Milestones and weights for real-world progress scoring. The weights sum to 1.0 for each task and are shared by all compared policies. A trial that completes the task but collides with a non-target object receives a progress score of 0.8.}
    \label{tab:real_robot_progress_settings}
    \begin{tabular}{@{}llp{0.55\textwidth}r@{}}
        \toprule
        Task & Step & Milestone & Weight \\
        \midrule
        \multirow{5}{*}{Place Object}
        & 1 & Approach the white block & 0.15 \\
        & 2 & Align the gripper orientation for grasping & 0.30 \\
        & 3 & Grasp the white block & 0.20 \\
        & 4 & Move the held block toward the gray bowl & 0.15 \\
        & 5 & Release the white block inside the gray bowl & 0.20 \\
        \cmidrule(lr){2-4}
        & & Total & 1.00 \\
        \midrule
        \multirow{5}{*}{Pack Box}
        & 1 & Approach the brown toy & 0.15 \\
        & 2 & Align the gripper orientation for grasping & 0.30 \\
        & 3 & Grasp the brown toy & 0.20 \\
        & 4 & Move the held toy toward the green basin & 0.15 \\
        & 5 & Release the brown toy inside the green basin & 0.20 \\
        \cmidrule(lr){2-4}
        & & Total & 1.00 \\
        \midrule
        \multirow{6}{*}{Stack Blocks}
        & 1 & Approach the white block & 0.10 \\
        & 2 & Align the gripper orientation for grasping & 0.20 \\
        & 3 & Grasp the white block & 0.20 \\
        & 4 & Move the held block toward the red block & 0.10 \\
        & 5 & Align the held block's orientation for placement & 0.20 \\
        & 6 & Release the white block stably on top of the red block & 0.20 \\
        \cmidrule(lr){2-4}
        & & Total & 1.00 \\
        \bottomrule
    \end{tabular}
\end{table*}

\noindent \textbf{Ravens.}
For each task setting and demonstration budget, we evaluate 100 episodes on the released initialization seeds \cite{shridhar2022cliport}. An episode ends when the task is completed or the action budget is exhausted. We report the mean task completion score, computed as $N^{-1}\sum_{e=1}^{N}R_e$, where $N=100$ and $R_e$ is the sum of the environment rewards over episode $e$. The benchmark awards partial credit based on matched object poses or occupancy of target regions, so incomplete multi-step tasks can receive nonzero scores. 

\noindent \textbf{LIBERO.}
Following the OpenVLA-OFT evaluation protocol \cite{kim2025fine}, each evaluation contains 50 episodes per task and 10 tasks per suite. Before executing policy actions, the environment advances for 10 no-op steps to let objects settle. The subsequent action limits are 220, 280, 300, and 520 steps for \textit{Spatial}, \textit{Object}, \textit{Goal}, and \textit{Long}, respectively. An episode is successful when all task-goal predicates are satisfied within the action limit; otherwise, it is counted as a failure. Following the official protocol, reported results are averaged over three random seeds, and the overall score is the arithmetic mean across the four suites.

\noindent \textbf{VLABench.}
We use the released scenario configurations for the five evaluation tracks \cite{zhang2025vlabench}. Rollouts follow the evaluator's task-specific step limits. We use the benchmark's discrete IS: an episode receives an intention score of 1 if the minimum distance between the end effector and the task-designated target object during the rollout is below the evaluator's distance threshold $\delta=0.15\,\mathrm{m}$, and 0 otherwise. IS therefore assesses whether the policy approaches the intended target. PS measures partial task completion by combining the number of target objects successfully grasped during the rollout with the number of task termination conditions satisfied at its end, divided by the total number of target objects and termination conditions. For each task, IS and PS are averaged over its evaluated episodes. For our summary, track scores are arithmetic means across tasks, and Avg. is the arithmetic mean across the five tracks.

\noindent \textbf{RoboTwin 2.0.}
Following TurboVLA's latest evaluation protocol \cite{xie2026turbovla}, we evaluate all 50 tasks under both Clean and Randomized settings, with 100 episodes per task in each setting. This gives 5,000 episodes per setting and 10,000 episodes per policy across both settings. Clean uses nominal environmental conditions, whereas Randomized varies background textures, lighting, tabletop clutter, and table height. Episode success is determined by the benchmark's task-specific success flag. We first compute each task's success rate, then average equally over the 50 tasks to obtain the Clean and Randomized scores. Avg. is the arithmetic mean of these two scores.

\noindent \textbf{Real-World Evaluation.}
OpenVLA-OFT and each Eq.Bot variant are evaluated for 10 trials on each of \textit{Place Object}, \textit{Pack Box}, and \textit{Stack Blocks}, yielding 30 trials per policy. We report both progress scores and full task success rates. All policies are assessed with the same task definitions and scoring rubric.

Progress scores are assigned from video replay using the milestones in Table~\ref{tab:real_robot_progress_settings}. Milestones are credited in the listed order, each at most once per trial, and previously earned credit is retained if a later stage fails. During execution, the robot and the manipulated object must avoid collisions with non-target objects. Grasping requires the target object to be secured by the gripper, and placement requires its release at the instructed destination. Milestones for grasp and placement orientation are judged from the visible gripper or held-object pose relative to the intended manipulation. \textit{Stack Blocks} includes a separate placement-orientation milestone because stable stacking requires more precise geometric alignment than placement into a container.

For task $t$, let $w_{t,k}$ be the incremental weight of milestone $k$, and let $K_{t,e}$ be the last consecutively completed milestone in trial $e$. The weights sum to 1 for each task. A trial that achieves the instructed final configuration but collides with a non-target object receives a progress score of 0.8 (80\%) and is counted as a failure. The per-trial score and reported task score are:
\begin{equation}
    \begin{aligned}
        s_{t,e} &=
        \begin{cases}
            0.8, & \text{goal achieved with a collision},\\
            \displaystyle\sum_{k=1}^{K_{t,e}}w_{t,k}, & \text{otherwise},
        \end{cases}\\
        P_t &= \frac{100}{N_t}\sum_{e=1}^{N_t}s_{t,e}.
    \end{aligned}
    \label{eq:real_robot_progress}
\end{equation}
where $N_t=10$ and the milestone sum is zero if no milestone is completed. In the scoring rule, a collision refers to contact with a non-target object. 

Full task success requires achieving the instructed final configuration without any collision with non-target objects. \textit{Place Object} requires the white block to be released inside the gray bowl; \textit{Pack Box} requires the brown toy to be released inside the green basin; and \textit{Stack Blocks} requires the white block to remain stably on top of the red block after release. For each task, the success rate is $100N_t^{-1}\sum_{e=1}^{N_t}y_{t,e}$, where $y_{t,e}=1$ only if the final task objective is achieved without such a collision, and 0 otherwise. The overall success rate is the mean across the three tasks.

\subsection*{C.4 Supplementary Simulation Results}

\begin{table*}[t]
    \centering
    \small
    \setlength{\tabcolsep}{3pt}
    \caption{OptNet-CNN with Transformer-based policies. LIBERO and RoboTwin 2.0 report success rates (\%); VLABench reports intention score (IS) and progress score (PS), both in percent. Avg. denotes the mean across suites, settings, or tracks, respectively. Emb. PT. indicates additional embodied pretraining beyond RoboTwin 2.0.}
    \label{tab:optnet_cnn_transformers}
    \begin{minipage}[t]{0.45\linewidth}
        \centering
        \textbf{(a) LIBERO}\par\smallskip
        \begin{tabular*}{\linewidth}{@{\extracolsep{\fill}}lccccc@{}}
            \toprule
            Method & Spatial & Object & Goal & Long & Avg. \\
            \midrule
            OpenVLA-OFT & 96.6 & 97.8 & 97.6 & 94.1 & 96.5 \\
            Eq.Bot (OptNet-CNN) & 97.4 & 98.2 & 97.8 & 94.6 & 97.0 \\
            \bottomrule
        \end{tabular*}
    \end{minipage}\hfill
    \begin{minipage}[t]{0.52\linewidth}
        \centering
        \textbf{(b) RoboTwin 2.0}\par\smallskip
        \begin{tabular*}{\linewidth}{@{\extracolsep{\fill}}lccccc@{}}
            \toprule
            Method & Emb.~PT. & Params~(B) & Clean & Rand. & Avg. \\
            \midrule
            TurboVLA & $\times$ & 0.4 & 87.52 & 87.58 & 87.55 \\
            Eq.Bot (OptNet-CNN) & $\times$ & 0.4 & 88.46 & 89.08 & 88.77 \\
            \bottomrule
        \end{tabular*}
    \end{minipage}
    \par\medskip
    \textbf{(c) VLABench}\par\smallskip
    \begin{tabular*}{\linewidth}{@{\extracolsep{\fill}}lcccccccccccc@{}}
        \toprule
        & \multicolumn{2}{c}{In-dist.} & \multicolumn{2}{c}{Category} & \multicolumn{2}{c}{Commonsense}
        & \multicolumn{2}{c}{Instruction} & \multicolumn{2}{c}{Texture} & \multicolumn{2}{c}{Avg.} \\
        \cmidrule(lr){2-3} \cmidrule(lr){4-5} \cmidrule(lr){6-7} \cmidrule(lr){8-9} \cmidrule(lr){10-11} \cmidrule(lr){12-13}
        Method & IS & PS & IS & PS & IS & PS & IS & PS & IS & PS & IS & PS \\
        \midrule
        $\pi_{0.5}$ & 73.2 & 59.6 & 50.0 & 32.9 & 61.5 & 43.3 & 60.2 & 27.2 & 60.1 & 48.9 & 61.0 & 42.4 \\
        Eq.Bot (OptNet-CNN) & 80.2 & 64.9 & 55.3 & 36.5 & 63.8 & 43.6 & 66.4 & 30.7 & 59.2 & 48.1 & 65.0 & 44.8 \\
        \bottomrule
    \end{tabular*}
    \vspace{-2ex}
\end{table*}

\begin{table}[t]
    \centering
    \small
    \setlength{\tabcolsep}{3pt}
    \caption{OptNet-CNN on Ravens. Task completion scores (\%) are reported for each demonstration budget. For CLIPORT and the other Eq.Bot variants, see Table~\ref{tab:CLIPORT_label}.}
    \label{tab:optnet_cnn_ravens}
    \begin{tabular*}{1.02\linewidth}{@{\extracolsep{\fill}}lrrrr@{}}
        \toprule
        & \multicolumn{4}{c}{Demonstrations per task} \\
        \cmidrule(lr){2-5}
        Task setting & \makebox[2em][c]{1} & \makebox[2em][c]{10}
        & \makebox[2em][c]{100} & \makebox[2em][c]{1000} \\
        \midrule
        Packing box pairs (\textcolor{blue}{seen}) & 50.8 & 74.9 & 93.7 & 99.1 \\
        Packing box pairs (\textcolor{red}{unseen}) & 45.8 & 67.3 & 74.2 & 75.3 \\
        Packing Google objects, seq.\ (\textcolor{blue}{seen}) & 14.7 & 56.0 & 87.9 & 96.7 \\
        Packing Google objects, seq.\ (\textcolor{red}{unseen}) & 30.8 & 45.3 & 62.7 & 72.3 \\
        Packing Google objects, group (\textcolor{blue}{seen}) & 47.8 & 64.9 & 87.2 & 96.7 \\
        Packing Google objects, group (\textcolor{red}{unseen}) & 50.7 & 59.2 & 77.2 & 76.5 \\
        Stacking block pyramid (\textcolor{blue}{seen}) & 21.5 & 52.3 & 92.0 & 99.5 \\
        Stacking block pyramid (\textcolor{red}{unseen}) & 14.1 & 27.8 & 30.2 & 34.0 \\
        Separating piles (\textcolor{blue}{seen}) & 51.7 & 58.7 & 94.9 & 98.0 \\
        Separating piles (\textcolor{red}{unseen}) & 43.8 & 56.2 & 75.3 & 82.1 \\
        Towers of Hanoi (\textcolor{blue}{seen}) & 53.8 & 83.9 & 99.7 & 100.0 \\
        Towers of Hanoi (\textcolor{red}{unseen}) & 41.9 & 85.3 & 99.7 & 100.0 \\
        Aligning rope & 22.8 & 58.6 & 91.5 & 93.1 \\
        Packing \textcolor{red}{unseen} shapes & 23.0 & 40.0 & 48.0 & 50.0 \\
        Assembling kits (\textcolor{blue}{seen}) & 12.8 & 17.8 & 47.0 & 52.4 \\
        Assembling kits (\textcolor{red}{unseen}) & 16.6 & 19.6 & 33.8 & 31.2 \\
        Putting blocks in bowls (\textcolor{blue}{seen}) & 30.1 & 51.2 & 98.8 & 100.0 \\
        Putting blocks in bowls (\textcolor{red}{unseen}) & 18.7 & 30.2 & 54.7 & 51.2 \\
        \midrule
        Avg. (CLIPORT) & 28.6 & 46.9 & 67.0 & 72.9 \\
        Avg. (OptNet-CNN) & 32.9 & 52.7 & 74.9 & 78.2 \\
        \bottomrule
    \end{tabular*}
    \vspace{-2ex}
\end{table}

\noindent \textbf{OptNet-CNN across Benchmarks.}
To complement the main comparisons, we additionally evaluate OptNet-CNN, a variant that uses a lightweight conventional CNN to score candidate frames. Tables~\ref{tab:optnet_cnn_transformers} and~\ref{tab:optnet_cnn_ravens} report results across the four simulation benchmarks.

On Ravens, OptNet-CNN improves CLIPORT's mean completion score across all four demo budgets. The largest mean gain is 7.95 percentage points at 100 demos, where the score on \textit{align-rope} rises from 62.4\% to 91.5\%. The gains thus extend beyond the few-demonstration setting. On LIBERO, OptNet-CNN improves OpenVLA-OFT across all four suites, raising average success from 96.5\% to 97.0\%. The largest gain is on \textit{Spatial} (96.6\% to 97.4\%), consistent with the benefit of the shared geometric frame for spatially conditioned action prediction.

On VLABench, OptNet-CNN improves $\pi_{0.5}$'s mean IS and PS from 61.0\% and 42.4\% to 65.0\% and 44.8\%, respectively. Both scores improve in four of the five tracks, with the largest gains on \textit{In-dist.} (7.0 and 5.3 percentage points). On \textit{Texture}, however, IS and PS decrease by 0.9 and 0.8 points, respectively, showing that the gains do not extend uniformly to appearance changes. On RoboTwin 2.0, OptNet-CNN raises TurboVLA's average success from 87.55\% to 88.77\%, improving both Clean and Randomized performance, with a larger gain under Randomized conditions (1.50 vs.\ 0.94 points).

Taken together, these results show that the same canonicalization interface can improve aggregate performance with a lightweight encoder across policies that predict spatial maps, Cartesian actions, or joint targets, while the gains vary by task and evaluation setting.

\begin{table}[t]
    \centering
    \setlength{\tabcolsep}{4pt}
    \caption{Ravens rotation group ablation with 100 demonstrations per task. Entries are completion scores (\%); $C_n$ denotes continuous planar rotations.}
    \label{tab:ablation_study_100}
    \resizebox{\linewidth}{!}{%
        \begin{tabular}{lcccccccc}
            \toprule
            & \multicolumn{4}{c}{pack \textcolor{blue}{seen} objects} & \multicolumn{4}{c}{assemble \textcolor{blue}{seen} kits} \\
            \cmidrule(lr){2-5} \cmidrule(lr){6-9}
            Method & $C_2$ & $C_4$ & $C_8$ & $C_n$ & $C_2$ & $C_4$ & $C_8$ & $C_n$ \\
            \midrule
            Eq.Bot (OptNet-CNN) & 85.2 & \textbf{87.9} & \underline{86.5} & 83.0 & 42.4 & \textbf{47.0} & 44.2 & \underline{45.3} \\
            Eq.Bot (OptNet-Res50) & 86.9 & \textbf{89.2} & 87.6 & \underline{87.8} & 42.5 & \textbf{48.2} & 46.0 & \underline{48.0} \\
            Eq.Bot (DirNet) & 87.7 & \textbf{89.4} & 87.8 & \underline{88.0} & 42.8 & \textbf{51.6} & 49.3 & \underline{50.4} \\
            \midrule
            & \multicolumn{4}{c}{pack \textcolor{red}{unseen} objects} & \multicolumn{4}{c}{assemble \textcolor{red}{unseen} kits} \\
            \cmidrule(lr){2-5} \cmidrule(lr){6-9}
            Method & $C_2$ & $C_4$ & $C_8$ & $C_n$ & $C_2$ & $C_4$ & $C_8$ & $C_n$ \\
            \midrule
            Eq.Bot (OptNet-CNN) & 58.1 & \underline{62.7} & \textbf{62.8} & 59.5 & 27.6 & \textbf{33.8} & \underline{30.2} & 29.7 \\
            Eq.Bot (OptNet-Res50) & 58.7 & \textbf{62.9} & 62.4 & \underline{62.8} & 28.8 & \underline{32.0} & 29.8 & \textbf{32.8} \\
            Eq.Bot (DirNet) & 60.1 & \textbf{63.3} & 62.5 & \underline{63.0} & 29.2 & \textbf{31.0} & \underline{30.4} & 30.2 \\
            \bottomrule
        \end{tabular}%
    }
\vspace{-2ex}
\end{table}

\begin{table*}[t]
\centering
\setlength{\tabcolsep}{4pt}
\caption{Per-trial progress scores on \textit{Place Object}, \textit{Pack Box}, and \textit{Stack Blocks}. Scores range from 0 to 1; Total is the sum over the ten trials. Bold marks the highest total for each task.}
\resizebox{0.65\linewidth}{!}{%
  \begin{tabular}{lccccccccccc}
        \toprule
        \multicolumn{12}{c}{\textit{Place Object}} \\
        \midrule
        \multirow{2}{*}{Method} & \multicolumn{10}{c}{Trial} & \multirow{2}{*}{Total} \\
        \cmidrule(lr){2-11}
        & \makebox[1.5em][c]{1} & \makebox[1.5em][c]{2} & \makebox[1.5em][c]{3} & \makebox[1.5em][c]{4} & \makebox[1.5em][c]{5} & \makebox[1.5em][c]{6} & \makebox[1.5em][c]{7} & \makebox[1.5em][c]{8} & \makebox[1.5em][c]{9} & \makebox[1.5em][c]{10} & \\
        \midrule
        OpenVLA-OFT & 1 & 1 & 1 & 1 & 1 & 1 & 0.45 & 0 & 0.45 & 0.45 & 7.35 \\
        \midrule
        Eq.Bot (OptNet-CNN) & 1 & 1 & 1 & 1 & 1 & 1 & 1 & 0.45 & 0.45 & 0.45 & 8.35 \\
        Eq.Bot (OptNet-Res50) & 1 & 1 & 1 & 1 & 1 & 1 & 1 & 0.80 & 0.80 & 0.80 & \textbf{9.40} \\
        Eq.Bot (DirNet) & 1 & 1 & 1 & 1 & 1 & 1 & 1 & 1 & 0.80 & 0.45 & 9.25 \\
        \midrule
        \multicolumn{12}{c}{\textit{Pack Box}} \\
        \midrule
        \multirow{2}{*}{Method} & \multicolumn{10}{c}{Trial} & \multirow{2}{*}{Total} \\
        \cmidrule(lr){2-11}
        & \makebox[1.5em][c]{1} & \makebox[1.5em][c]{2} & \makebox[1.5em][c]{3} & \makebox[1.5em][c]{4} & \makebox[1.5em][c]{5} & \makebox[1.5em][c]{6} & \makebox[1.5em][c]{7} & \makebox[1.5em][c]{8} & \makebox[1.5em][c]{9} & \makebox[1.5em][c]{10} & \\
        \midrule
        OpenVLA-OFT & 1 & 1 & 1 & 1 & 1 & 1 & 1 & 1 & 0.45 & 0.80 & 9.25 \\
        \midrule
        Eq.Bot (OptNet-CNN) & 1 & 1 & 1 & 1 & 1 & 1 & 1 & 1 & 0.80 & 1 & 9.80 \\
        Eq.Bot (OptNet-Res50) & 1 & 1 & 1 & 1 & 1 & 1 & 1 & 1 & 0.80 & 1 & 9.80 \\
        Eq.Bot (DirNet) & 1 & 1 & 1 & 1 & 1 & 1 & 1 & 1 & 1 & 1 & \textbf{10.00} \\
        \midrule
        \multicolumn{12}{c}{\textit{Stack Blocks}} \\
        \midrule
        \multirow{2}{*}{Method} & \multicolumn{10}{c}{Trial} & \multirow{2}{*}{Total} \\
        \cmidrule(lr){2-11}
        & \makebox[1.5em][c]{1} & \makebox[1.5em][c]{2} & \makebox[1.5em][c]{3} & \makebox[1.5em][c]{4} & \makebox[1.5em][c]{5} & \makebox[1.5em][c]{6} & \makebox[1.5em][c]{7} & \makebox[1.5em][c]{8} & \makebox[1.5em][c]{9} & \makebox[1.5em][c]{10} & \\
        \midrule
        OpenVLA-OFT & 1 & 1 & 1 & 1 & 1 & 0.60 & 0.50 & 0.30 & 0.30 & 0.30 & 7.00 \\
        \midrule
        Eq.Bot (OptNet-CNN) & 1 & 1 & 1 & 1 & 1 & 1 & 0.60 & 0.30 & 0.30 & 0.30 & 7.50 \\
        Eq.Bot (OptNet-Res50) & 1 & 1 & 1 & 1 & 1 & 1 & 0.80 & 1 & 0.30 & 0.50 & 8.60 \\
        Eq.Bot (DirNet) & 1 & 1 & 1 & 1 & 1 & 1 & 1 & 0.80 & 0.30 & 0.60 & \textbf{8.70} \\
        \bottomrule
  \end{tabular}}
\label{tab:progress_place_object}
\end{table*}

\noindent \textbf{Rotation Groups with 100 Demonstrations.}
Table~\ref{tab:ablation_study_100} repeats the comparison in Table~\ref{tab:ablation_study} with 100 demos per task. $C_4$ gives the highest completion score in 10 of the 12 task-canonicalizer comparisons. The exceptions at 100 demos are OptNet-CNN on unseen-object packing, for which $C_8$ leads by 0.1 point, and OptNet-Res50 on unseen-kit assembly, for which $C_n$ leads by 0.8 points. This pattern supports $C_4$ as a default at both budgets: quarter turns broaden coverage over $C_2$ while avoiding the angular interpolation required by finer rotations on a compatible pixel grid.

\subsection*{C.5 Real-World Trial Results}

Tables~\ref{tab:progress_place_object} list the progress scores of all ten trials for each policy on \textit{Place Object}, \textit{Pack Box}, and \textit{Stack Blocks}. Entries are the per-trial scores $s_{t,e}\in[0,1]$ defined in Appendix~C.3, including the collision rule. Total is the sum over ten trials.

The trial records reveal differences that binary success alone does not capture. On \textit{Place Object}, CNN and ResNet50 both complete seven of ten trials, but the three remaining trials score 0.45 each for CNN and 0.80 each for ResNet50. Their progress totals are therefore 8.35 and 9.40, despite identical success rates. A score of 0.80 may reflect partial progress or task completion penalized for a collision, so it is not counted as full success. On \textit{Stack Blocks}, the totals increase from 7.00 for OpenVLA-OFT to 8.60 and 8.70 for ResNet50 and GNet, respectively. These records complement the main comparison by exposing variation in the progress achieved across trials.

\section*{D. Limitations and Future Works}

Eq.Bot couples observation canonicalization with action recovery through a shared estimated frame, allowing both operations to remain outside the base policy. The current implementation focuses on planar rotations and uses $C_4$ by default. The ablations with 10 and 100 demonstrations support this choice, although finer rotations perform better in some settings (Tables~\ref{tab:ablation_study} and~\ref{tab:ablation_study_100}). General $SE(3)$ transformations remain outside the evaluated scope: out-of-plane motion and changes in camera viewpoint can alter perspective and visibility in ways that planar image warps cannot reproduce. A natural extension is to estimate the shared frame from 3D scene geometry and use calibrated mappings between camera, workspace, and robot coordinates for observation alignment and action recovery.

The equivariance guarantee relies on consistent frame estimation and compatible group actions on observations and policy outputs. In practice, occlusion and scene symmetries can make the canonical orientation ambiguous, while image resampling can perturb the aligned observation. Small changes in the input may therefore cause the estimated frame to switch abruptly and disrupt action prediction. Future work could explicitly model orientation ambiguity and improve temporal consistency in frame selection. Measuring frame consistency and action equivariance error on paired transformed observations, alongside task performance, would help quantify how closely the implemented system satisfies the theoretical conditions.

Finally, Eq.Bot evaluates the base policy once per prediction, but frame estimation and the coordinate mappings still incur overhead. Table~\ref{tab:inference_time} shows that this overhead varies across canonicalizers and base policies. Future work could reduce the cost of frame scoring and action recovery and evaluate the resulting trade-off among latency, geometric consistency, and task performance.

\end{document}